\documentclass[11pt]{article}

\usepackage{acl}

\usepackage{times}
\usepackage{latexsym}
\usepackage{amsmath}
\usepackage{color}
\usepackage[T1]{fontenc}
\usepackage[utf8]{inputenc}

\usepackage{microtype}

\usepackage{inconsolata}

\usepackage{graphicx}

\title{Dual Perspectives in Emotion Attribution: A Generator-Interpreter Framework for Cross-Cultural Analysis of Emotion in LLMs}

\author{Aizirek Turdubaeva \\
  KAIST / Daejeon, Republic of Korea \\
  \texttt{aizturdubaeva@kaist.ac.kr} \\\And
  Uichin Lee \\
  KAIST / Daejeon, Republic of Korea \\ 
  \texttt{uclee@kaist.ac.kr} \\}

\definecolor{bcolor}{rgb}     {1.0,0.0,0.0} 

\definecolor{bcolor}{rgb}     {0.0,1.0,0.0} 

\definecolor{acolor}{rgb}     {0.0,0.0,1.0}

\definecolor{bcolor}{rgb}     {1.0,0.0,1.0}

\begin{document}
\maketitle
\begin{abstract}
%Cross-cultural systems have long been central in human-computer interactions. In this tradition, 
Large language models (LLMs) are increasingly used in cross-cultural systems to understand and adapt to human emotions, which are shaped by cultural norms of expression and interpretation. However, prior work on emotion attribution has focused mainly on interpretation, overlooking the cultural background of emotion generators. This assumption of universality neglects variation in how emotions are expressed and perceived across nations. To address this gap, we propose a Generator-Interpreter framework that captures dual perspectives of emotion attribution by considering both expression and interpretation. We systematically evaluate six LLMs on an emotion attribution task using data from 15 countries. Our analysis reveals that performance variations depend on the emotion type and cultural context. Generator-interpreter alignment effects are present; the generator’s country of origin has a stronger impact on performance. We call for culturally sensitive emotion modeling in LLM-based systems to improve robustness and fairness in emotion understanding across diverse cultural contexts.

% We call for culturally sensitive approaches in LLM-powered communication and social-emotional support applications to foster inclusive interaction design and user experience.

\end{abstract}

\section{Introduction}

Cultural differences in customs, social values, and everyday practices play an integral role in determining human behavior. Emotional experiences are subjective and strongly tied to sociocultural background and accumulated past experiences~\cite{milkowski_personal_2021}. They reflect internal states and influence how individuals perceive external stimuli~\cite{izard_emotion_2009} and inform behavior upon them~\cite{lerner_emotion_2015}. Consequently, emotions are not universal in how they are experienced, expressed, or interpreted.

Psychological theories have shown that \emph{display rules}~\cite{ekman_repertoire_1969, barrett_experience_2007} regulate how individuals in different societies express or suppress emotions, while \emph{attribution theory}~\cite{heider_psychology_2013} explains how observers judge or classify those expressions based on cultural expectations.
As a result, the same emotional event from one country may be judged very differently by interpreters in another. Emotion attribution is thus inherently cultural as both the act of expression (generation) and the act of judgment (interpretation) depend on sociocultural norms~\cite{barrett_experience_2007, mesquita_cultural_1992, matsumoto_mapping_2008}. 

In this study, we introduce a framework that considers the dual nature of emotion and accounts for the cultural background of both: (1) generation, which reflects cultural influences on how emotions are expressed (display norms), and (2) interpretation, which reflects the cultural lens through which emotions are judged (attribution).

\begin{figure}[t]
\centering
\includegraphics[width=0.9\linewidth]{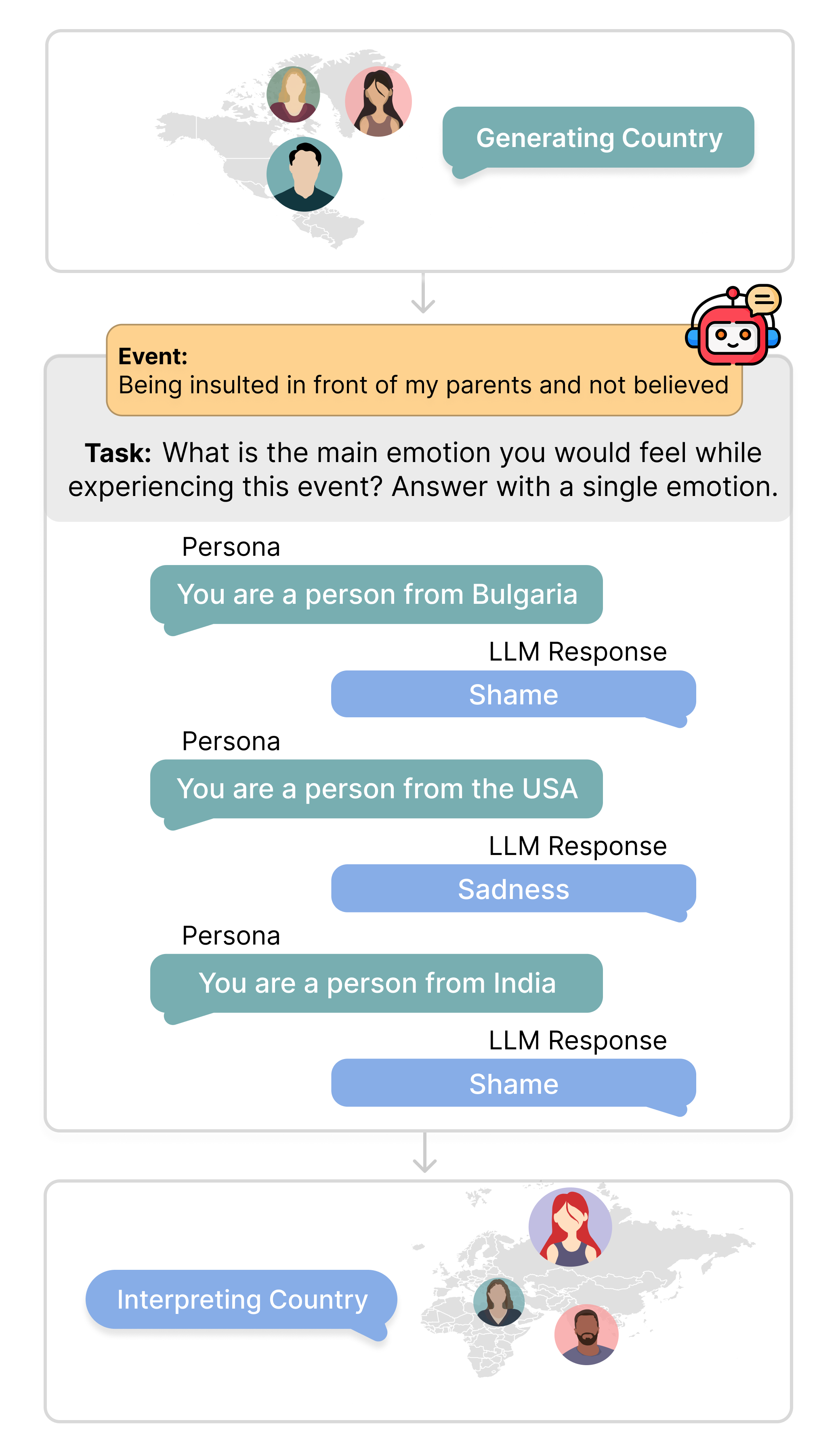}
\caption{Cross-cultural emotion attribution. An event labeled as anger is interpreted as shame or sadness.}
% framework. Emotional stories generated in one country may be interpreted differently when judged by interpreters from other cultural contexts.}
\label{fig:intro-framework}
\end{figure}

The need for such a framework grows with the global deployment of large language models (LLMs) across various tasks in cross-cultural systems~\cite{pang_understanding_2025}, making it crucial to ensure that they reflect cultural variations in emotion.
In particular, as multilingual tools like globally deployed conversational agents become more widespread, the risk of cultural misalignment in emotion attribution becomes more pronounced~\cite{havaldar_multilingual_2023, belay_culemo_2025}.
LLMs are increasingly used for social-emotional support applications~\cite{jo_understanding_2023}, such as mental health assistants~\cite{li_systematic_2023, kim_mindfuldiary_2024, seo_chacha_2024, xu_mental-llm_2024, yuan_improving_2025, ma_evaluating_2024}, educational tutors~\cite{bassner_iris_2024, kumar_guiding_2024, zhang_mathemyths_2024}, customer service support~\cite{das_swain_ai_2025}, and social robots~\cite{kim_understanding_2024, kang_nadine_2024}.
These applications demand emotional sensitivity to provide effective and empathetic social interaction support~\cite{cuadra_illusion_2024, breazeal_emotion_2003, fazzi_dont_2025}. 

However, current LLMs have been reported to be biased towards mainstream culture while ignoring others. The model's generative capacity is tightly linked to the training data, which appears to be disproportionately Western and Anglocentric~\cite{havaldar_multilingual_2023}. 
Prior work showed that LLMs reproduce cultural and gender stereotypes~\cite{plaza-del-arco_angry_2024, wan_kelly_2023, kotek_gender_2023}, fail to fully capture cultural variations associated with emotion, and predominantly reflect the cultural values of the Western world~\cite{plaza-del-arco_divine_2024, palta_fork_2023, naous_having_2024}.
%The consequences of such a built-in bias are of growing concern.
Cultural misalignment might undermine the emotional support these technologies are designed to provide.   

Despite longstanding research in emotion analysis, most existing studies examined only the interpreter’s side---how a model or person interprets emotions in text. This neglects the cultural background of the generators, the people producing those emotional texts. In other words, prior work has largely assumed that emotion recognition is a one-way process, overlooking how cross-cultural dynamics between the generator and interpreter may shape emotion attribution.

To the best of our knowledge, no systematic study has examined Generator x Interpreter country pairings on emotion attribution. 
To bridge this gap, we use the ISEAR dataset (5,250 events across 15 countries) and evaluate six well-known large language models, comparing within-country vs. cross-country attribution accuracy. 
We treat a generator as the country where an emotional event originated (e.g., a Swiss participant writing about a shame experience), and an interpreter as the nationality persona classifying the emotion (e.g., ``classify this as if you were an Indian annotator''). 
This study allows us to answer the following research questions:

\textit{\textbf{RQ1}: How do nationality personas differ in emotion attribution across nations?}

\textit{\textbf{RQ2}: How does generator-interpreter alignment (same country vs. different countries) affect attribution accuracy?} 

%\vspace{5}
Key contributions of this work include:  
\begin{itemize}
    \item We argue that for cross-cultural LLM systems, there is a need for the Generator x Interpreter evaluation framework, explicitly modeling cultural origin and target in emotion attribution.
    \item We conduct systematic evaluation with six LLMs across 15 nationality personas on the ISEAR dataset, uncovering cross-cultural differences and biases in emotion attribution.\footnote{Code and data are available at \url{https://anonymous.4open.science/r/llm-emotion-recognition-8354/}}
\end{itemize}

\section{Background and Related Work}

We review (1) basic psychological theories of emotion generation and recognition across cultural contexts, (2) prior findings on cultural and social biases in LLMs, and (3) emotion analysis in the natural language processing field.

\subsection{Theories of Emotion Generation and Recognition}
In the study of emotions, the long-standing universalist perspective argues that emotions are biologically hardwired and expressed in invariant ways across cultures. However, more recent scholarship has increasingly emphasized the constructionist view, which highlights that emotions are shaped through cultural, contextual, and social factors~\cite{barrett_constructionist_2024, barrett_theory_2017}. From this standpoint, it becomes crucial to account for contextual variability: anger may be expressed or suppressed depending on cultural display rules, and even seemingly universal emotions such as happiness can be conceptualized differently across societies~\cite{lu_culture_2004}. 
Building on this perspective, the appraisal theory of emotion~\cite{scherer_appraisal_2001} underscores that emotional reactions depend on individuals’ interpretation of events in relation to their goals and values, which differ across cultural contexts, while attribution theory~\cite{heider_psychology_2013} explains how observers judge and classify those expressions through culturally grounded expectations.
Furthermore, theories of emotion regulation~\cite{slovak_designing_2023} and display rules~\cite{ekman_repertoire_1969, matsumoto_cultural_1990} stress that people recognize and manage emotional expressions in line with social and cultural norms, amplifying, muting, or suppressing them to maintain appropriateness. 
Taken together, these theories highlight that both the generation and interpretation of emotions are culturally mediated. Consequently, when the locus of emotion generation and the locus of interpretation belong to different cultural contexts, misunderstandings are more likely to arise, and such cross-cultural asymmetries may amplify biases or distortions in how emotions are perceived and represented.

\subsection{Cultural and Social Biases}

Prior studies revealed that LLMs exhibit systematic cultural and social biases, often privileging Western norms while misrepresenting or stereotyping others.
LLMs often default to Western or American liberal-democratic values when probed on moral reasoning tasks~\cite{fraser_does_2022, abdulhai_moral_2024}. Survey-style probing with frameworks like Hofstede’s cultural dimensions and the World Values Survey confirms that models reproduce Western-aligned responses even in non-Western cultural contexts~\cite{masoud_cultural_2025}. 
Language Models capture some geo-diverse facts (e.g., Chinese bridal dress color~\cite{nguyen_extracting_2023}, culinary customs~\cite{palta_fork_2023}, culturally specific time expressions~\cite{shwartz_good_2022}), but often oversimplify or distort non-Western norms.
They also reproduce social stereotypes, e.g., portraying women as ``warm'' and men as ``role models'' in reference letters~\cite{wan_kelly_2023}, or associating anger with men and sadness with women~\cite{plaza-del-arco_angry_2024}. 
Other studies showed systematic privileging of Western-Christian contexts, while misrepresenting other cultures with stereotypes of poverty and traditionalism~\cite{naous_having_2024}, and disproportionately stigmatizing Judaism and Islam~\cite{plaza-del-arco_divine_2024}.
To the best of our knowledge, most existing studies have primarily focused on interpretation, examining how LLMs classify cultural knowledge, values, or emotions. However, prior work has not addressed the interaction between generation and interpretation across different cultural contexts. In particular, while users or language models may generate content reflecting one cultural setting, the interpretation or evaluation of that content often takes place from another. This generator-interpreter heterogeneity raises open questions about how cultural misalignment manifests when the loci of content generation and interpretation differ, and whether such cross-cultural asymmetries amplify biases or distortions in LLM outputs.

%Taken together, designing systems that can generate, recognize, and adapt emotional expressions across cultural boundaries is essential for equitable participation, social support, and inclusive user experiences.

\subsection{Emotion Attribution Research in NLP}
% you may want to briefly talk about what research was done on emotion attribution - cultural stuff was discussed in previous subsection. And then you talk about existing models didn't consider generator/interpretators simultaneously. 

% As LLMs increasingly mediate interpersonal communication, recent research has begun to examine their emotional sensitivity across cultural contexts. \citet{havaldar_multilingual_2023} show that multilingual LLMs tend to reproduce Western-centric norms even when responding in non-English languages, whereas~\citet{belay_culemo_2025} demonstrate that emotion conceptualizations vary substantially across languages and cultures. 

Emotion analysis is a central aspect of communication and has become a rapidly growing field in natural language processing~\cite{plaza-del-arco_emotion_2024}. Most prior work relies on either discrete (categorical) or dimensional emotional models. The discrete model represents emotions as a set of distinct categories, exemplified by Ekman's basic emotions and Plutchik's wheel of emotions. The dimensional model conceptualizes emotions as interconnected and represented along continuous axes, such as arousal and valence. Several surveys provide broad overviews of the emotion analysis field, covering datasets, models, detection approaches, metrics, applications, challenges and opportunities~\cite{acheampong_textbased_2020, saxena_emotion_2020, tomar_unimodal_2023, murthy_review_2021}. Existing datasets predominantly focus on emotion recognition in conversational settings, while more recent ones incorporate social media posts, news headlines, poems, and customer reviews. However, a recent survey by~\citet{plaza-del-arco_emotion_2024} finds that most studies omit demographic and cultural information about data creators or annotators, revealing a critical gap in existing work. Since uncovering biases and social stereotypes in LLMs requires considering both emotion and demographics, our study is the first to account for the cultural background of emotion \textit{generators} and \textit{interpreters}.

\section{Methods}
\subsection{Dataset}
We used the ISEAR (International Survey on Emotion Antecedents and Reactions) dataset, which was collected during the 1990s~\cite{scherer_evidence_1994}.\footnote{Dataset URL: \url{https://www.unige.ch/cisa/index.php/download_file/view/395/296/}}The dataset contains 7,666 self-reported emotional events from nearly 3,000 student participants, representing 37 countries across five continents. The original data from 16 countries were transcribed and translated into English. The dataset covers seven emotion categories: joy, fear, anger, sadness, disgust, shame, and guilt. For each emotional event, participants provided: (1) a personal description of the situation, typically written in their native language, (2) an appraisal of the situation and how they reacted, (3) demographic information, including but not limited to gender, religion, and nationality. While the ISEAR dataset is fully translated into English, it retains the cultural origin of each emotional narrative (i.e., the generator’s country). Thus, our study does not aim to capture linguistic nuances per se, but rather the cultural framing of emotional experiences. By holding language constant, we focus on how interpreter personas attribute emotions differently depending on the generator’s cultural background. The ISEAR dataset was used due to rich generator-level demographic metadata, which are largely absent from most contemporary emotion datasets~\cite{plaza-del-arco_emotion_2024}.

\subsection{Data Cleaning and Exclusion Criteria}
\label{sec:methods_cleaning}
The dataset was decoded according to the ISEAR Questionnaire \& Codebook~\cite{noauthor_research_2015}. To ensure data quality, several cleaning steps were applied. First, null or invalid responses were removed, such as ``Nothing,'' ``No response,'' ``Not applicable,'' and ``Can’t remember that feeling.'' Second, we excluded non-serious or cross-referenced responses that lacked unique situational content. %Examples include statements such as ``Same as in anger'' and ``The same for guilt would apply.'' 
A complete list of excluded responses is provided in Table~\ref{tab:excluded-responses} in Appendix. Finally, text normalization was performed by removing special characters (e.g., ``á'') and correcting minor inconsistencies such as bracket usage. After cleaning, the final dataset consisted of 7,472 valid emotional event reports, with 192 invalid entries removed.   
To balance the uneven number of emotions per country, we randomly sampled 50 events per emotion category, yielding 350 events per country. Norway was excluded due to insufficient data. This produced a final dataset of 5,250 (15 countries × 350) events for this study.

\subsection{Model Selection}
We evaluated six recent state-of-the-art LLMs known for their strong performance across various tasks, such as GPT-4o~\cite{openai_gpt-4_2024}, Gemini~\cite{team_gemini_2024}, Claude~\cite{noauthor_anthropics_nodate}, Mistral~\cite{jiang_mistral_2023}, Qwen2~\cite{qwen_qwen25_2024}, and DeepSeek~\cite{deepseek-ai_deepseek-v3_2024}. We aimed to include both open-source and proprietary models, ranging across different sizes and development origins. To explore cultural influences in model outputs, we tried to balance Western-developed (e.g., GPT-4o, Gemini, Claude) and Eastern-developed (e.g., Qwen2, DeepSeek) models in our study. Details of each model are provided in Table~\ref{tab:model_parameters} in Appendix.

\subsection{Emotion Attribution Task}
We adopted the emotion-attribution task designed by~\citet{plaza-del-arco_angry_2024}, which provides a structured framework for analyzing cultural biases in emotion recognition. Given an emotional event from the ISEAR dataset and a specified country persona, the model is asked to determine which emotion the persona would most likely experience. All experiments were conducted in a zero-shot manner, without providing any examples. Models were prompted with a temperature of 0.0 to ensure consistency in terms of response variability across models and a maximum response length of 256 tokens. We used the following task prompt:
\textit{``What is the main emotion you would feel while experiencing this event \{event\}? Answer with a single emotion and omit explanations. Emotion:''}

\subsection{Prompt Design and Evaluation}
Recent research in NLP explored persona-based system prompts~\cite{shao_character-llm_2023} to surface intrinsic biases in large language models~\cite{gupta_bias_2023, deshpande_toxicity_2023, aher_using_2023}.
% Accordingly, we employ a persona-based method to examine any potential biases and stereotypes in emotion attribution across models.}
We assigned 15 country personas using a tailored persona template introduced by Gupta et al.~\cite{gupta_bias_2023}. Specifically, each persona corresponds to the country from which the ISEAR event was originally reported. This setup would allow us to analyze any patterns or insights in how models interpret emotions based on the persona's cultural background, as well as potential mismatches between the interpreting persona and the origin of the emotional event. The final persona prompt is shown below, with only the country name varying across personas:   

\textit{``You are a person from Zambia. Your responses should closely mirror the knowledge and abilities of this persona. What is the main emotion you would feel while experiencing this event, ``When I had a serious argument with a dear person''? You have to answer with a single emotion out of the following: anger, fear, sadness, joy, disgust, guilt, or shame. Omit explanations. Emotion:.''}

With 350 events from each of 15 countries (5,250 in total), we prompted each event using all 15 country personas across 6 different LLMs. This resulted in a total of 472,500 emotion attributions (5,250 x 15 x 6). We adopted a bootstrap resampling approach~\cite{efron_bootstrap_1992}. Specifically, we performed random sampling with replacement, repeating this procedure 100 times to obtain more robust performance estimates. 

For each resampled set, we computed macro-level metrics, including accuracy, precision, recall, and F1 score. Among these metrics, accuracy is reported as our primary evaluation measure in the main text, as it provides the most interpretable summary of overall model performance in this task setting. Detailed results are included in the Appendix for completeness.
% for precision, recall, and F1 score 

\section{Results}
\label{sec:results}

\subsection{Performance Across Nationality Personas}

\begin{figure*}[t]
\centering
\includegraphics[width=0.80\textwidth]{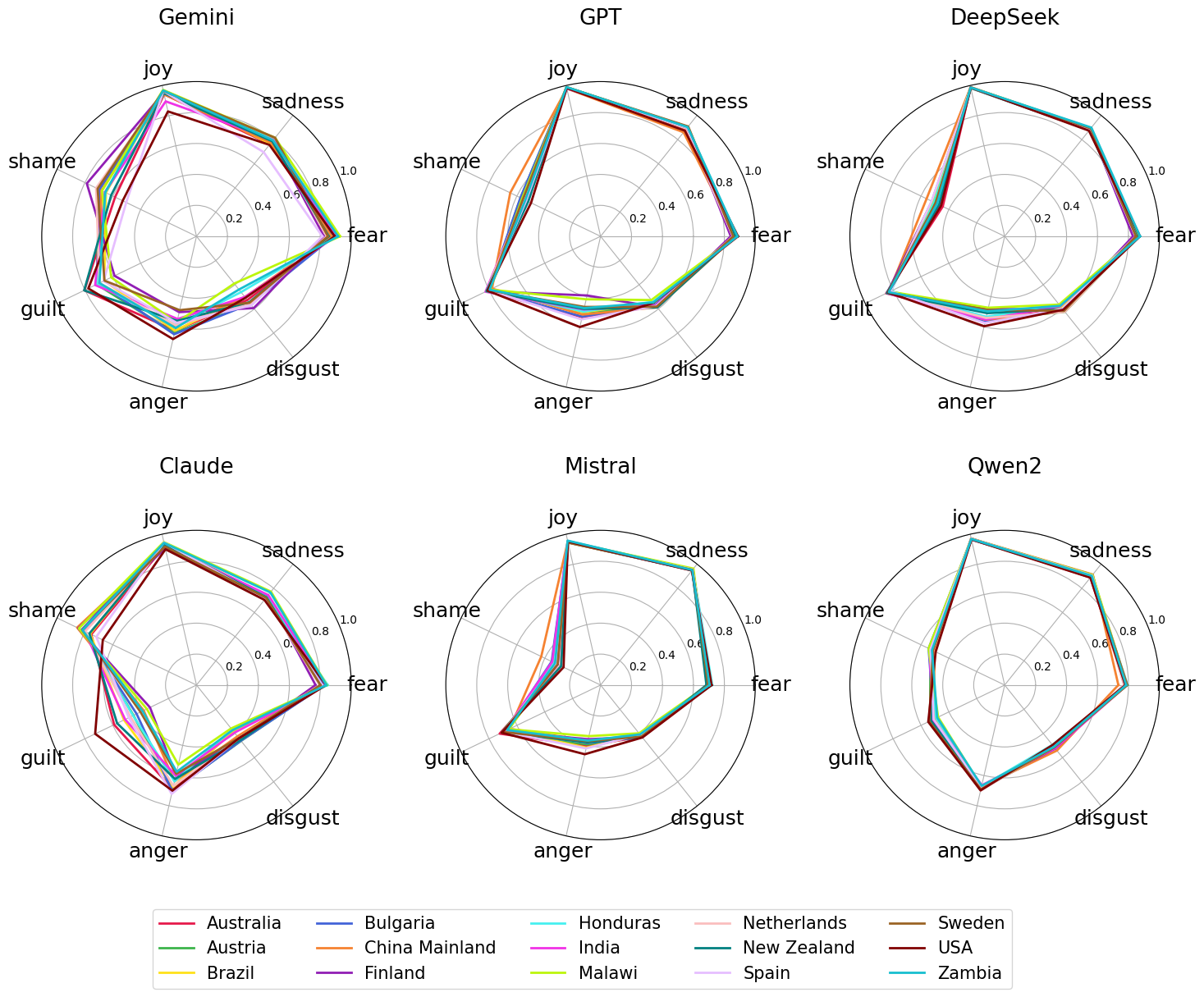}
\caption{Emotion attribution accuracy for each model, averaged across all personas.}
\label{fig:llm-results}
\end{figure*}

% \begin{figure}[t]
%     \centering
%     \includegraphics[width=\linewidth]{figures/radar plots.png}
%     \caption{Emotion attribution accuracy for each model, averaged across all personas}
%     \label{fig:llm-results}
% \end{figure}

We first examine how interpreter personas differ in overall performance when attributing emotions. Accuracy across personas is relatively stable, with most models scoring between 0.70 and 0.76 (Table~\ref{tab:persona_performance}). Although GPT-4o and DeepSeek achieve the highest accuracy (0.75--0.76) and Mistral the lowest (0.63), accuracy alone hides substantial differences. F1 scores show far greater divergence across models and personas, because they reflect precision-recall tradeoffs that expose systematic biases in how emotions are attributed.

%\subsubsection{Which emotions are easy or difficult to classify across nationality personas?}

We observe clear asymmetries across emotions (Table~\ref{tab:emotion_performance}). Joy emerges as the most universal, with both precision and recall high and F1 consistently above 0.93 across all models. Fear also shows robust performance, though recall is slightly lower than precision, placing it just behind joy. By contrast, sadness is often over-predicted: Mistral, for example, achieves a recall of 0.95 but at the cost of very low precision, while other models achieve more balanced performance (F1:0.67--0.73). Shame and guilt remain difficult to detect, with F1 rarely exceeding 0.65, likely due to systematic misattribution to sadness. Anger shows divergent behavior across models, whereas some under-recognize it almost entirely (e.g., Mistral) and others, such as GPT-4o, Gemini, and DeepSeek, recover higher recall and achieve F1 around 0.70. Finally, disgust is the most challenging emotion overall, with F1 consistently the lowest across models, ranging only from 0.37 to 0.65.

%\subsubsection{What are misclassification patterns?} 

\begin{figure}[h]
    \centering
    \includegraphics[width=0.85\linewidth]{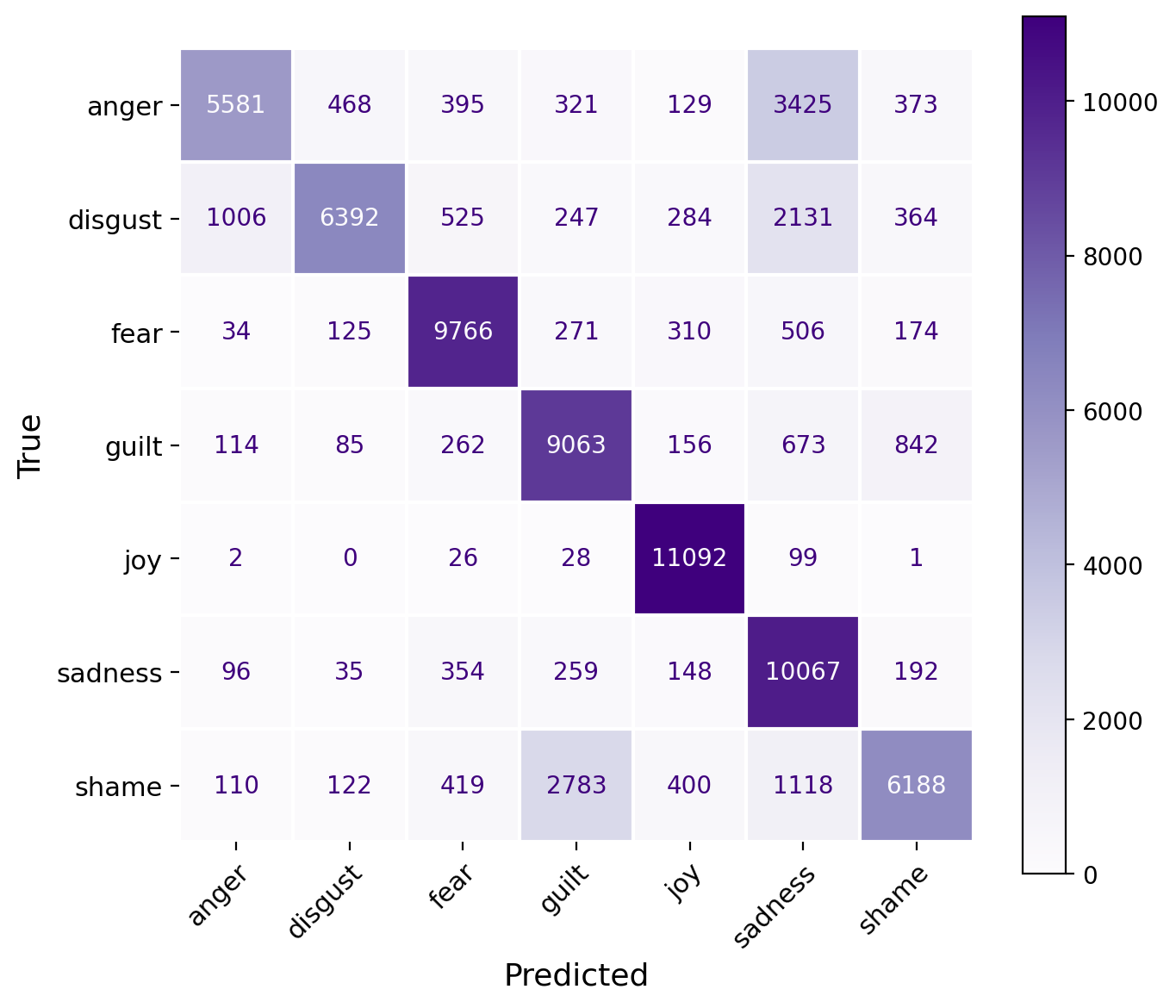}
    \caption{Confusion matrix aggregated across personas in the GPT-4o model.}
    \label{fig:gpt-confusion-matrix}
\end{figure}

We next analyze which emotions are most frequently misclassified (Figure~\ref{fig:gpt-confusion-matrix}). The strongest bidirectional errors occur between \emph{guilt} and \emph{shame}, which are frequently confused with each other across all models and personas. A second frequent pattern is the confusion between \emph{anger} and \emph{disgust}, with both categories often substituted for one another. Beyond these pairs, we also observe a one-directional bias in which multiple emotions collapse into \emph{sadness}. \emph{Anger, disgust, and shame} are commonly mislabeled as \emph{sadness}, leading to sadness dominating attribution patterns in models such as Mistral and Qwen2. Together, these results show that misclassifications are not random but instead follow systematic patterns shaped by semantic and affective proximity between emotions.

These misclassification patterns closely align with established psychological and cultural research. Shame and guilt, both self-conscious emotions, are often intertwined in collectivist societies (e.g., China, India), where shame serves as a tool of social cohesion and control. In contrast, individualist societies (e.g., Sweden, US) place greater emphasis on personal guilt as a marker of individual responsibility~\cite{kitayama_culture_1995, wong_cultural_2007, eid_norms_2001}. Disgust, meanwhile, can be expressed in both physical forms (e.g., food, hygiene) and moral forms (e.g., injustice, betrayal), each shaped by cultural norms. Moral disgust is frequently articulated in language resembling anger, while self-directed disgust may overlap with sadness or shame, leading models to misclassify it as other negative emotions~\cite{herz_stealing_2013, lee_maggots_2013}. Large language models appear to mirror these dynamics: they assign a higher proportion of shame responses to Chinese personas, while more accurately classifying guilt for individualistic societies, thereby reproducing and reinforcing existing cultural patterns.

\begin{table*}[t]
\centering
\footnotesize
\setlength{\tabcolsep}{3pt}
\renewcommand{\arraystretch}{1.05}
\begin{tabular}{l*{16}{c}}
\hline
& \multicolumn{16}{c}{\textbf{Interpreter}} \\
\hline
\textbf{Generator}
& AU & AT & BR & BG & CN & FI & HN & IN & MW & NL & NZ & ES & SE & US & ZM & \textbf{Avg} \\
\hline
AU & \textbf{0.74} & 0.74 & 0.71 & 0.75 & 0.73 & 0.71 & 0.71 & 0.72 & 0.71 & 0.74 & 0.73 & 0.74 & 0.74 & 0.74 & 0.73 & 0.73 \\
AT & 0.81 & \textbf{0.81} & 0.79 & 0.81 & 0.80 & 0.77 & 0.80 & 0.79 & 0.78 & 0.81 & 0.79 & 0.81 & 0.79 & 0.81 & 0.79 & 0.80 \\
BR & 0.78 & 0.78 & \textbf{0.75} & 0.78 & 0.78 & 0.76 & 0.77 & 0.77 & 0.76 & 0.77 & 0.77 & 0.78 & 0.77 & 0.78 & 0.77 & 0.77 \\
BG & 0.77 & 0.76 & 0.76 & \textbf{0.76} & 0.77 & 0.75 & 0.75 & 0.76 & 0.76 & 0.76 & 0.75 & 0.75 & 0.76 & 0.75 & 0.75 & 0.76 \\
CN & \textcolor{red}{0.63} & \textcolor{red}{0.63} & \textcolor{red}{0.62} & \textcolor{red}{0.63} & \textbf{\textcolor{red}{0.67}}
 & \textcolor{red}{0.60} & \textcolor{red}{0.61} & \textcolor{red}{0.62} & \textcolor{red}{0.59} & \textcolor{red}{0.62} & \textcolor{red}{0.62} & \textcolor{red}{0.62} & \textcolor{red}{0.62} & \textcolor{red}{0.65} & \textcolor{red}{0.63} & \textcolor{red}{0.62} \\
FI & 0.73 & 0.73 & 0.73 & 0.74 & 0.74 & \textbf{0.71} & 0.72 & 0.73 & 0.71 & 0.72 & 0.73 & 0.73 & 0.73 & 0.73 & 0.73 & 0.73 \\
HN & 0.79 & 0.79 & 0.79 & 0.79 & 0.80 & 0.78 & \textbf{0.78} & 0.79 & 0.76 & 0.79 & 0.79 & 0.80 & 0.79 & 0.79 & 0.77 & 0.79\\
IN & 0.77 & 0.79 & 0.77 & 0.79 & 0.79 & 0.77 & 0.77 & \textbf{0.79} & 0.77 & 0.79 & 0.78 & 0.79 & 0.78 & 0.77 & 0.77 & 0.78\\
MW & 0.80 & 0.79 & 0.80 & 0.81 & 0.80 & 0.79 & 0.81 & 0.83 & \textbf{0.80} & 0.79 & 0.79 & 0.80 & 0.79 & 0.80 & 0.81 & 0.80\\
NL & 0.74 & 0.73 & 0.73 & 0.74 & 0.75 & 0.71 & 0.72 & 0.73 & 0.71 & \textbf{0.73} & 0.73 & 0.74 & 0.73 & 0.74 & 0.72 & 0.73 \\
NZ & 0.70 & 0.72 & 0.71 & 0.72 & 0.73 & 0.70 & 0.72 & 0.72 & 0.69 & 0.70 & \textbf{0.71} & 0.72 & 0.71 & 0.73 & 0.71 & 0.71 \\
ES & 0.76 & 0.75 & 0.76 & 0.76 & 0.75 & 0.74 & 0.75 & 0.76 & 0.75 & 0.76 & 0.75 & \textbf{0.76} & 0.75 & 0.77 & 0.73 & 0.75\\
SE & 0.70 & 0.70 & 0.69 & 0.70 & 0.70 & 0.69 & 0.69 & 0.69 & 0.69 & 0.69 & 0.69 & 0.69 & \textbf{0.70} & 0.68 & 0.70 & 0.69\\
US & 0.72 & 0.73 & 0.72 & 0.75 & 0.73 & 0.72 & 0.72 & 0.74 & 0.73 & 0.74 & 0.73 & 0.73 & 0.73 & \textbf{0.72} & 0.72 & 0.73\\
ZM & 0.70 & 0.70 & 0.67 & 0.69 & 0.69 & 0.67 & 0.68 & 0.69 & 0.65 & 0.69 & 0.69 & 0.69 & 0.69 & 0.70 & \textbf{0.68} & 0.69 \\
\hline
\textbf{Avg} & 0.74 & 0.74 & 0.73 & 0.75 & 0.75 & 0.72 & 0.73 & 0.74 & 0.72 & 0.74 & 0.74 & 0.75 & 0.74 & 0.75 & 0.74 & 0.74\\
\hline
\end{tabular}
\caption{GPT-4o Model: Generator-Interpreter Accuracy Results. Red indicates the lowest performance value per interpreter column. Note that AU = Australia, AT = Austria, BR = Brazil, BG = Bulgaria, CN = China, FI = Finland, HN = Honduras, IN = India, MW = Malawi, NL = Netherlands, NZ = New Zealand, ES = Spain, SE = Sweden, US = USA, ZM = Zambia.}
\label{tab:gpt4o_alignment}
\end{table*}

\subsection{Generator-Interpreter Alignment}
\label{sec:generator-interpreter-alignment}

We first examine whether accuracy is higher when the generator and interpreter share the same country persona. Some alignment effects are visible across models, but they are not consistent in all cases. For example, in GPT-4o (Table~\ref{tab:gpt4o_alignment}), Austria (0.81), China (0.67), and Malawi (0.80) achieve higher accuracy when the generator and interpreter are aligned. DeepSeek shows a similar pattern, with Austria and Malawi both reaching 0.81 (Table~\ref{tab:deepseek_alignment}). By contrast, Mistral (Table~\ref{tab:mistral_alignment}) shows only modest alignment effects, and in some cases, self-alignment does not outperform cross-persona interpretations. For instance, China achieves only 0.45 in self-alignment, which is lower than when interpreted by Brazil (0.48), and Zambia records 0.59, below Australia’s score when interpreting Zambian events (0.61). These results suggest that while alignment can improve performance in certain contexts, it does not universally guarantee higher accuracy.

However, a clearer trend across all models is that events generated by some countries are systematically harder to interpret than others. Events generated by China and Zambia are consistently difficult for all interpreters, with average accuracy dropping to 0.62 in GPT-4o, 0.55 in Qwen2 (Table~\ref{tab:qwen_alignment}), and as low as 0.46 in Mistral (Table~\ref{tab:mistral_alignment}). Even the strongest models, such as DeepSeek (Table~\ref{tab:deepseek_alignment}), remain below global averages for Chinese-generated events. By contrast, narratives from Austria, Brazil, Spain, and Malawi are interpreted far more reliably, including by the Chinese persona. These findings highlight that the generator side strongly influences and shapes emotion attribution difficulty. For example, when Chinese personas interpret Chinese events, accuracy improves compared to cross-persona cases. Yet their scores still fall short of those achieved on most Western-generated events.

We then examine whether some countries are ``better interpreters'' than others. Most interpreter personas perform quite consistently, with only slight differences across countries. Yet the USA, Austria, Spain, Bulgaria, and Australia interpreter personas achieve systematically higher accuracy across generators and models. For instance, Bulgaria, the USA, and Spain reach around 0.75 in GPT-4o (Table~\ref{tab:gpt4o_alignment}), while Austria and the USA achieve 0.71 in Claude (Table~\ref{tab:claude_alignment}), and DeepSeek (Table~\ref{tab:deepseek_alignment}) shows 0.75 for Bulgaria, the USA, Spain, and Austria. 
By contrast, many personas in Qwen2 (Table~\ref{tab:qwen_alignment}) cluster around 0.70, reflecting the overall narrow spread of results. Still, even the strongest interpreters experience substantial declines when classifying events generated by China and Zambia, underscoring that performance is also determined by the generation side, with some produced events that are hard for other personas to interpret.

\section{Discussion}

Recent studies have highlighted that LLMs reflect persistent societal biases and stereotypes in emotion attribution. One line of research found that models reproduce gendered stereotypes, overwhelmingly predicting \textit{sadness} for women and \textit{anger} for men~\cite{plaza-del-arco_angry_2024}. Another study revealed religious biases, showing that major Western religions tend to be represented with more nuance, while Eastern religions such as Judaism and Islam are often stigmatized~\cite{plaza-del-arco_divine_2024}. While our work similarly draws on the ISEAR dataset, it introduces a new perspective by examining nationality-based emotion attribution and the important role of generator-interpreter alignment.

Our study reveals that while nationality personas exhibit relatively stable overall accuracy in emotion attribution, systematic biases remain evident. Joy and fear are consistently well-recognized across models, whereas shame, guilt, and disgust persist as challenging categories, often collapsing into sadness. Accuracy is shaped not only by the interpreter but also by generator-interpreter alignment. Events originating from certain countries, particularly China, are systematically harder for all personas to interpret, though in certain cases, performance improves when the generator and interpreter share the same cultural background. These findings underscore the importance of considering the cultural origin of generated narratives, while also revealing that current interpretations often rely on superficial lexical cues rather than culturally grounded emotion understanding~\cite{belay_culemo_2025}. %a deeper understanding of cultural context and distinct ways emotions are expressed across nations~\cite{belay_culemo_2025}.

The global reach of LLMs has enabled their integration across diverse cultures, contexts, and user profiles. However, for effective generator-interpreter alignment in the LLM domain, it is crucial to explicitly account for the cultural background of the message generator to ensure accurate and contextually appropriate interpretation. In real-world LLMs serving diverse users worldwide, systems should leverage signals, such as user location, native language, language fluency, or textual style, to infer the interpreter's origin. Incorporating this cultural context into prompts or persona settings enables better alignment between the generated message and its interpreter, allowing LLMs to produce socially aware and culturally sensitive responses tailored to different interpreters. 

Currently, many systems remain dominated by Western-centric data, which perpetuates biases in model outputs and weakens performance on underrepresented cultures~\cite{fraser_does_2022, abdulhai_moral_2024}. To address this, it is essential to build and collect more diverse data that reflects linguistic differences and represents low-resource countries and languages. Strengthening the generator’s side with such data will improve multicultural systems’ ability to understand users’ backgrounds and languages, promoting respect and inclusivity in cross-cultural communication.

Furthermore, our results have important implications for cross-cultural systems that understand and adapt to human emotions~\cite{duan_bridging_2021, choi_toward_2023, zhang_facilitating_2022}. Communication in the real world often unfolds in multicultural, multiparty interactions where one message may be interpreted by multiple individuals from diverse cultural backgrounds, or where multiple generators and interpreters interact simultaneously. Our findings emphasize that effective alignment requires consideration not only of the interpreter’s perspective but also of the generator’s cultural origin, which significantly influences how emotions are expressed and understood. Alignment between generator and interpreter is typically strongest when both share the same cultural background; however, in multicultural settings where multiple generators and interpreters from diverse cultures interact, misalignment can lead to biases and reduced accuracy. This complexity raises important questions about how alignment should be understood and managed when cultural backgrounds vary widely among participants. Of course, auditing models with \emph{Generator $\times$ Interpreter} evaluations helps. Additionally, designers should explicitly incorporate cultural backgrounds and persona profiles into system prompts to tailor interpretations and additionally offer emotional reasoning~\cite{yang_towards_2023}. However, since LLMs cannot inherently know users’ cultural contexts without explicit input, it is critical for systems to encourage or facilitate users specifying their cultural identity or for designers to build interaction mechanisms that surface this information thoughtfully. By foregrounding cultural information in prompts and interfaces, applications can better navigate the complexities of multicultural interactions, reduce bias, and foster clearer, more empathetic communication.

\section{Conclusion}
We introduced a Generator-Interpreter framework for cross-cultural emotion attribution and conducted a systematic evaluation of six large language models across 15 country personas using the ISEAR dataset. Across models, overall accuracy was generally stable, yet we observed consistent and interpretable asymmetries by emotion and culture: \emph{joy} and \emph{fear} were reliably recognized, whereas \emph{sadness} tended to be over-predicted and \emph{shame}, \emph{guilt}, and \emph{disgust} remained challenging. While generator-interpreter alignment may help improve performance in some cases, although not consistently, we found that the \emph{generator’s country of origin exerts a stronger influence} on attribution difficulty than the interpreter alone. These findings indicate that emotion attribution in LLMs is shaped by both expression (generation) and judgment (interpretation), and that overlooking the generator side obscures systematic errors.
Our results underscore the need for \emph{culture-awareness by design} when we build social-emotional support systems and cross-cultural support agents. In practice, this entails (i) capturing or inferring the cultural origin of the input (e.g., user-provided context, language, and style signals), (ii) conditioning interpretation on culturally grounded personas or profiles, and (iii) auditing models with \emph{Generator $\times$ Interpreter} evaluations rather than interpreter-only benchmarks. Such practices can reduce misattribution, support more empathetic interactions, and foster inclusive experiences for globally diverse users.
 
\section{Limitations}
There are several limitations of this work. 
The ISEAR dataset covers only seven basic emotion categories based on Ekman’s theory~\cite{ekman_repertoire_1969}, which may fail to capture fine-grained, nuanced emotion categories.  %While his work provides a framework for basic emotions, it wasn’t intended to offer a comprehensive or context-sensitive taxonomy of emotional experience. Emotions are far more nuanced, and many complex or culturally specific emotional states are excluded from this framework. 
%Relying solely on these broad categories may oversimplify the emotional landscape and 
%This oversimplification may limit the depth of insights in cross-cultural emotion studies. 
%Further, only 15 countries from the ISEAR dataset were included in our analysis; thus, 
Further, our findings may not generalize to underrepresented or excluded regions from the dataset.
%There are certain limitations we acknowledge in our study. 
%Emotions are highly tied to cultural context and the language associated with it.
Due to the limited dataset, our study relied exclusively on English-only prompts and translations, which may not fully capture culturally nuanced emotional expressions as reported in the prior studies~\cite{havaldar_multilingual_2023, bareis_english_2024}. %Prior research by Barreiß et al.~\cite{bareis_english_2024} has shown that English prompts outperform target-language prompts in zero-shot settings.
%Similarly, 
%Havaldar et al.~\cite{havaldar_multilingual_2023} demonstrated that prompting in a target language results in less culturally-aware emotional response than communicating in English with a prefix stating the nationality.
Future research should investigate multilingual prompts and datasets with diverse persona settings, such as gender and religion~\cite{plaza-del-arco_divine_2024}, to better disentangle cultural variation from translation effects.

\bibliography{zotero}

\appendix

%\section{Appendix}
\label{sec:appendix}


\section{Appendix}
This appendix provides supporting details for our experimental setup and evaluation. Section~\ref{sec:app_models} summarizes the specifications of the six large language models used in our study (release date, openness, parameter size when available, and license). Section~\ref{sec:app_cleaning} documents the ISEAR dataset cleaning process, including concrete examples of excluded responses (null/invalid entries and cross-referenced answers). Section~\ref{sec:app_results} reports the full emotion attribution results, including persona-level performance across 15 nationality interpreters, emotion-level precision/recall/F1 across the seven target emotions, and generator--interpreter accuracy matrices for each model. Section~\ref{sec:app_diagnostics} provides deeper diagnostic analyses, including confusion matrices for representative generator countries (high- and low-accuracy cases) and illustrative qualitative examples of Chinese-origin events that reveal systematic cross-persona misattribution patterns, such as shifts between shame and guilt, attenuation of anger, and misrecognition of culturally embedded disgust. Finally, Sections~\ref{sec:app_structured} and~\ref{sec:app_generator_spec} report additional evaluations on structured output constraints and explicit disclosure of generator nationality.

\section{Model Parameters}\label{sec:app_models}
Model specifications, including release dates, parameter sizes, open-source availability, and license types, are presented in Table \ref{tab:model_parameters}.

\begin{table*}[t]
\centering
\setlength{\tabcolsep}{10pt}
\small
\begin{tabular}{l c c c c}
\hline
\textbf{Model} & \textbf{Release} & \textbf{Size} & \textbf{Open} & \textbf{License} \\
\hline
gpt-4o-2024-05-13         & May 2024 & --   & X & Proprietary \\
gemini-pro-1.5           & Apr 2024 & --   & X & Proprietary \\
claude-3.5-sonnet        & Oct 2024 & --   & X & Proprietary \\
mistral-7b-instruct-v0.3 & May 2024 & 7.3B & O & Apache 2.0 \\
qwen-2.5-7b-instruct     & Oct 2024 & 7B   & O & Apache 2.0 \\
deepseek-v3              & Dec 2024 & 671B & O & MIT \\
\hline
\end{tabular}
\caption{Model specifications including release dates, parameter sizes, open-source availability, and license types.}
\label{tab:model_parameters}
\end{table*}

\section{ISEAR Dataset Cleaning}\label{sec:app_cleaning}

As described in Section~\ref{sec:methods_cleaning}, we applied a cleaning process to ensure data quality. Appendix Table~\ref{tab:excluded-responses} provides concrete examples of responses that were excluded.  We distinguish two main categories: (1) \textbf{Null/Invalid}, where participants gave no meaningful content (e.g., “Nothing,” “No response,” “Can’t remember that feeling”), and 
(2) \textbf{Cross-Referenced}, where participants did not provide a unique event but instead referred to another response category (e.g., “Same as in anger,” “The same for guilt would apply”). This table illustrates the list of entries we have excluded from our analysis.

\begin{table*}[h]
\centering
\footnotesize
\setlength{\tabcolsep}{6pt}
\renewcommand{\arraystretch}{1.05}
\begin{tabular}{p{0.18\textwidth} p{0.78\textwidth}}
\hline
\textbf{Category} & \textbf{Examples of excluded responses} \\
\hline
\textbf{Null / Invalid} &
“Nothing.”, “None.”, “Not applicable.”, “NO RESPONSE.”, “Not applicable to myself.”, 
“Doesn't apply.”, “Can't remember having had this feeling.”, “Can't remember that feeling.”, 
“Can't remember any episode of disgust.”, “Can't remember any such feelings now.”, 
“I cannot recall feeling guilty in the near past. REMARK - SUBJECT MAY NOT HAVE TAKEN THE EXERCISE VERY SERIOUSLY.”, 
“Blank.”, “I can't remember ever feeling shame.”, “Not included on questionnaire.”, 
“Cannot recall the emotion with any force.”, “Haven't felt shame for ages.”, 
“Cannot remember such a situation.”, “DO NOT REMEMBER.”, 
“I do not recall being scared in the near past.”, “Can't remember exact example.”, 
“I think I have hardly had a situation where I felt disgusted. Perhaps once, when I saw a big spider.”, 
“Can think of no time that I have felt fear.”, 
“When I considered writing a load of crap for this emotion.”, 
“Sadness – … I can't remember a concrete example.”, 
“Never really had the experience just yet – N/A.”, 
“I can't remember being truly angry. Usually it's just disappointment that I feel.”, 
“Disgust is a feeling that I have never experienced with people but only with repulsive animals (e.g., snakes on T.V.).”, 
“I cannot remember a situation in which I felt ashamed… knocked down my own glass.”, “[ Do not know.]”, “[ No response.]”, “[ Can not think of anything just now.]”, “[ Can not think of any situation.]”, “[ Can not remember.]”, 
“[ Cannot recall any incident when I felt shame.]”, “[ No answer as I see very little difference between shame and guilt. Therefore see answer for "shame".]”, “[ Do not remember any situation of that kind.]”, “[ Can not think of anything.]”, “[ I can not remember anything in particular. What I can remember is feelings of having done or said something which then had negative consequences.]”, 
“[ No description.]”, “[ I have not felt this emotion.]”, “[ Never felt the emotion.]”, 
“[ I do not remember any event when I felt guilty.]”, “[ I have never felt this emotion.]”, “[ I can not recall one.]”, 
“[ The previous incident holds good here also.]”, “[ ....has not really happened.]”, “[ Never experienced.]”, “[ Never.]”, 
“[ Sorry, I was never ashamed about anything in my life.]”, “[ Do not remember any incident.]”, 
“[ Normally I do not feel disgusted.]”, “[ I have not felt this emotion yet.]”, “[ I have not felt this emotion in my life.]”, 
“[ I have felt shame but am unable to remember any particular incident.]”, “[ I do not recall one here either.]”, “[ I was not, I am not, I will never ever feel ashamed of myself for something I have done.]”, “[ There have been very few instances of disgust, as far as I can remember they are not important or worth mentioning.]”, “[ Never felt guilty, as I have never done anything which could make me feel guilty.]”, “[ Honestly, I have never felt disgust at any situation in my life.]”, “[ I can positively say that I have never done anything that made me feel guilty.]”, “[ There are many instances which are all equally irritating.]”\\
\hline
\textbf{Cross-Referenced} &
“The same as in ‘shame’.”, “The same event described under ‘shame’.”, “The same situation as with sadness. I was afraid about what my parents would further do to destroy my relationship.”, 
“The same as for SHAME and DISGUST. I was asked to resign my sorority which I did because I was ashamed.”, 
“The same as in SHAME.”, “The same for guilt would apply.”, “Same situation as before – having my car stolen.”, 
“As for disgust.”, “I do not seem to feel emotion such as shame, guilt or fear – far out.”, “Robbery mentioned under sadness.”, “[ Same as in anger.]”, “[ Same as above – friends who torture animals.]”, “[ The same as in shame.]”, “[ The same as in anger.]”, “[ The same as in guilt.]”, “The same event described under FEAR” \\
\hline
\end{tabular}
\caption{Examples of responses excluded during ISEAR dataset cleaning.}
\label{tab:excluded-responses}
\end{table*}

\section{Emotion Attribution Evaluation Results}\label{sec:app_results}

This appendix lists the full tables corresponding to the evaluation results reported in Section~\ref{sec:results}. Table~\ref{tab:persona_performance} reports accuracy and F1-scores across 15 nationality personas, 
and Table~\ref{tab:emotion_performance} shows precision, recall, and F1-scores for the seven target emotions. The remaining tables provide detailed Generator–Interpreter Accuracy matrices for 
Gemini (Table~\ref{tab:gemini_alignment}), 
Mistral (Table~\ref{tab:mistral_alignment}), 
DeepSeek (Table~\ref{tab:deepseek_alignment}), 
Qwen2 (Table~\ref{tab:qwen_alignment}), 
and Claude (Table~\ref{tab:claude_alignment}). 
Here, \textit{Generator} indicates the country where an emotional event originated, and \textit{Interpreter} indicates the nationality persona classifying the emotion.

\begin{table*}[h]
\centering
\small
\setlength{\tabcolsep}{5pt}
\renewcommand{\arraystretch}{1.15}
\begin{tabular}{lcccccccccccc}
\hline
 & \multicolumn{2}{c}{\textbf{Mistral}} 
 & \multicolumn{2}{c}{\textbf{Qwen2}} 
 & \multicolumn{2}{c}{\textbf{Gemini}} 
 & \multicolumn{2}{c}{\textbf{GPT-4o}} 
 & \multicolumn{2}{c}{\textbf{DeepSeek}} 
 & \multicolumn{2}{c}{\textbf{Claude}} \\
\cline{2-13}
\textbf{Persona}
 & \textbf{Acc.} & \textbf{F1}
 & \textbf{Acc.} & \textbf{F1}
 & \textbf{Acc.} & \textbf{F1}
 & \textbf{Acc.} & \textbf{F1}
 & \textbf{Acc.} & \textbf{F1}
 & \textbf{Acc.} & \textbf{F1} \\
\hline
Australia   & 0.63 & 0.06 & 0.70 & 0.14 & 0.73 & 0.14 & \textbf{0.74} & 0.25 & \textbf{0.74} & \textbf{0.27} & 0.70 & 0.04 \\
Austria     & 0.63 & 0.05 & 0.70 & 0.15 & \textbf{0.75} & 0.16 & 0.74 & 0.23 & \textbf{0.75} & \textbf{0.25} & 0.70 & 0.06 \\
Brazil      & 0.64 & 0.06 & 0.70 & 0.14 & 0.72 & 0.11 & 0.73 & 0.22 & \textbf{0.74} & \textbf{0.23} & 0.69 & 0.05 \\
Bulgaria    & 0.63 & 0.06 & 0.70 & 0.15 &\textbf{0.75} & 0.18 & \textbf{0.75} & \textbf{0.31} & \textbf{0.75} & 0.25 & 0.70 & 0.07\\
Sweden      & 0.63 & 0.05 & 0.70 & 0.15 & 0.72 & 0.13 & \textbf{0.74} & 0.25 & \textbf{0.74} & \textbf{0.30} & 0.67 & 0.04 \\
Finland     & 0.63 & 0.05 & 0.70 & 0.16 & 0.72 & 0.14 & 0.73 & \textbf{0.27} & \textbf{0.74} & 0.26 & 0.66 & 0.04 \\
Honduras    & 0.63 & 0.06 & 0.69 & 0.16 & 0.73 & 0.20 & 0.73 & 0.24 & \textbf{0.74} & \textbf{0.28} & 0.68 & 0.04 \\
India       & 0.63 & 0.06 & 0.70 & 0.16 & 0.72 & 0.17 & \textbf{0.74} & 0.29 & \textbf{0.74} & \textbf{0.30} & 0.69 & 0.08 \\
Malawi      & 0.62 & 0.06 & 0.70 & 0.15 & 0.72 & \textbf{0.31} & 0.72 & \textbf{0.31} & \textbf{0.73} & 0.27 & 0.66 & 0.05 \\
Zambia      & 0.63 & 0.06 & 0.69 & 0.16 & 0.72 & 0.24 & 0.73 & 0.23 & \textbf{0.74} & \textbf{0.26} & 0.67 & 0.05 \\
USA         & 0.64 & 0.05 & 0.70 & 0.12 & 0.71 & 0.12 &  0.74 & \textbf{0.29} &\textbf{0.75} & 0.23 & 0.71 & 0.04 \\
Spain       & 0.64 & 0.06 & 0.70 & 0.16 & 0.68 & 0.08 & 0.74 & \textbf{0.27} & \textbf{0.75} & 0.24 & 0.69 & 0.05 \\
China Mainland  & 0.64 & 0.06 & 0.69 & 0.17 & 0.72 & 0.17 & 0.75 & \textbf{0.31} & \textbf{0.76} & \textbf{0.31} & 0.68 & 0.06 \\
New Zealand  & 0.63 & 0.05 & 0.70 & 0.14 & 0.73 & 0.12 & \textbf{0.74} & 0.22 & 0.73 & \textbf{0.23} & 0.70 & 0.05 \\
Netherlands   & 0.63 & 0.05 & 0.70 & 0.15 & \textbf{0.74} & 0.14 & \textbf{0.74} & \textbf{0.27} & \textbf{0.74} & 0.25 & 0.69 & 0.04 \\
\hline
\end{tabular}
\caption{Emotion attribution performance (accuracy and F1-score) of six LLMs across 15 interpreter personas. The highest accuracy and F1-score for each persona are shown in bold.}
\label{tab:persona_performance}
\end{table*}

\begin{table*}[h]
\centering
% \footnotesize
\small
\setlength{\tabcolsep}{3pt}
\renewcommand{\arraystretch}{1.05}
\begin{tabular}{lcccccccccccccccccc}
\hline
& \multicolumn{3}{c}{\textbf{Mistral}}
& \multicolumn{3}{c}{\textbf{Qwen2}}
& \multicolumn{3}{c}{\textbf{Gemini}}
& \multicolumn{3}{c}{\textbf{GPT-4o}}
& \multicolumn{3}{c}{\textbf{DeepSeek}}
& \multicolumn{3}{c}{\textbf{Claude}} \\
\hline
\textbf{Emotion} & \textbf{P} & \textbf{R} & \textbf{F1}
& \textbf{P} & \textbf{R} & \textbf{F1}
& \textbf{P} & \textbf{R} & \textbf{F1}
& \textbf{P} & \textbf{R} & \textbf{F1}
& \textbf{P} & \textbf{R} & \textbf{F1}
& \textbf{P} & \textbf{R} & \textbf{F1} \\
\hline
fear     & 0.91 & 0.70 & 0.79 & 0.86 & 0.78 & 0.82 & 0.75 & 0.87 & 0.81 & 0.83 & 0.87 & 0.85 & 0.83 & 0.86 & 0.84 & 0.81 & 0.82 & 0.81 \\
sadness  & 0.39 & 0.95 & 0.55 & 0.53 & 0.90 & 0.67 & 0.68 & 0.78 & 0.73 & 0.56 & 0.89 & 0.69 & 0.58 & 0.89 & 0.70 & 0.71 & 0.73 & 0.72 \\
joy      & 0.89 & 0.95 & 0.92 & 0.88 & 0.97 & 0.92 & 0.93 & 0.95 & 0.94 & 0.89 & 0.99 & 0.93 & 0.88 & 0.99 & 0.93 & 0.93 & 0.93 & 0.93 \\
shame    & 0.73 & 0.32 & 0.44 & 0.70 & 0.53 & 0.60 & 0.64 & 0.65 & 0.65 & 0.76 & 0.55 & 0.64 & 0.75 & 0.49 & 0.60 & 0.49 & 0.79 & 0.61 \\
guilt    & 0.62 & 0.69 & 0.65 & 0.77 & 0.52 & 0.62 & 0.72 & 0.70 & 0.71 & 0.70 & 0.81 & 0.75 & 0.67 & 0.83 & 0.74 & 0.78 & 0.47 & 0.59 \\
anger    & 0.76 & 0.39 & 0.51 & 0.62 & 0.68 & 0.65 & 0.70 & 0.59 & 0.64 & 0.80 & 0.50 & 0.61 & 0.78 & 0.53 & 0.63 & 0.63 & 0.64 & 0.64 \\
disgust  & 0.96 & 0.42 & 0.58 & 0.84 & 0.51 & 0.64 & 0.86 & 0.52 & 0.65 & 0.88 & 0.57 & 0.69 & 0.88 & 0.60 & 0.71 & 0.90 & 0.42 & 0.57 \\
\hline
\end{tabular}
\caption{Emotion attribution performance (P: precision, R: recall, F1) of six LLMs across seven target emotions.}
\label{tab:emotion_performance}
\end{table*}

% \section{Generator-Interpreter Matrix}

\begin{table*}[h]
\centering
\small
\setlength{\tabcolsep}{3pt}
\renewcommand{\arraystretch}{1.05}
\begin{tabular}{l*{16}{c}}
\hline
& \multicolumn{16}{c}{\textbf{Interpreter}} \\
\cline{2-17}
\textbf{Generator}
& AU & AT & BR & BG & CN & FI & HN & IN & MW & NL & NZ & ES & SE & US & ZM & \textbf{Avg} \\
\hline
AU & \textbf{0.73} & 0.77 & 0.73 & 0.77 & 0.72 & 0.75 & 0.73 & 0.72 & 0.71 & 0.76 & 0.74 & 0.71 & 0.75 & 0.73 & 0.72 & 0.74\\
AT & 0.83 & \textbf{0.84} & 0.81 & 0.83 & 0.79 & 0.78 & 0.79 & 0.79 & 0.76 & 0.82 & 0.81 & 0.78 & 0.81 & 0.82 & 0.79 & 0.80 \\
BR & 0.76 & 0.77 & \textbf{0.75} & 0.77 & 0.74 & 0.74 & 0.75 & 0.76 & 0.74 & 0.77 & 0.76 & 0.72 & 0.76 & 0.74 & 0.75 & 0.75\\
BG & 0.76 & 0.76 & 0.74 & \textbf{0.79} & 0.74 & 0.74 & 0.74 & 0.74 & 0.72 & 0.75 & 0.75 & 0.67 & 0.74 & 0.74 & 0.74 & 0.74 \\
CN & 0.61 & 0.59 & 0.58 & 0.61 & \textbf{0.63} & 0.56 & 0.60 & 0.61 & 0.58 & 0.58 & 0.59 & 0.50 & 0.55 & 0.58 & 0.62 & 0.59\\
FI & 0.73 & 0.77 & 0.73 & 0.77 & 0.75 & \textbf{0.74} & 0.72 & 0.74 & 0.73 & 0.75 & 0.73 & 0.70 & 0.74 & 0.72 & 0.73 & 0.74\\
HN & 0.75 & 0.77 & 0.76 & 0.77 & 0.75 & 0.75 & \textbf{0.77} & 0.76 & 0.75 & 0.75 & 0.76 & 0.72 & 0.75 & 0.72 & 0.77 & 0.75\\
IN & 0.75 & 0.77 & 0.75 & 0.78 & 0.73 & 0.77 & 0.77 & \textbf{0.74} & 0.75 & 0.77 & 0.77 & 0.67 & 0.76 & 0.71 & 0.76 & 0.75\\
MW & 0.74 & 0.75 & 0.73 & 0.77 & 0.74 & 0.72 & 0.76 & 0.71 & \textbf{0.73} & 0.73 & 0.73 & 0.69 & 0.72 & 0.73 & 0.74 & 0.73\\
NL & 0.73 & 0.75 & 0.74 & 0.75 & 0.73 & 0.75 & 0.73 & 0.71 & 0.71 & \textbf{0.75} & 0.73 & 0.70 & 0.73 & 0.73 & 0.72 & 0.73 \\
NZ & 0.71 & 0.72 & 0.70 & 0.69 & 0.69 & 0.67 & 0.71 & 0.70 & 0.67 & 0.70 & \textbf{0.70} & 0.69 & 0.67 & 0.69 & 0.70 & 0.69\\
ES & 0.74 & 0.75 & 0.72 & 0.75 & 0.73 & 0.72 & 0.73 & 0.73 & 0.70 & 0.73 & 0.75 & \textbf{0.69} & 0.72 & 0.71 & 0.71 & 0.73\\
SE & 0.68 & 0.73 & 0.70 & 0.73 & 0.70 & 0.70 & 0.70 & 0.68 & 0.68 & 0.73 & 0.69 & 0.65 & \textbf{0.70} & 0.65 & 0.70 & 0.69\\
US & 0.73 & 0.74 & 0.74 & 0.75 & 0.73 & 0.73 & 0.73 & 0.73 & 0.72 & 0.76 & 0.75 & 0.71 & 0.73 & \textbf{0.71} & 0.75 & 0.73\\
ZM & 0.69 & 0.70 & 0.68 & 0.69 & 0.68 & 0.67 & 0.68 & 0.66 & 0.67 & 0.71 & 0.69 & 0.66 & 0.68 & 0.69 & \textbf{0.69} & 0.68\\
\hline
\textbf{Avg} & 0.73 & 0.75 & 0.72 & 0.75 & 0.72 & 0.72 & 0.73 & 0.72 & 0.71 & 0.74 & 0.73 & 0.68 & 0.72 & 0.71 & 0.73 & 0.72 \\ 
\hline
\end{tabular}
\caption{Gemini Model: Generator-Interpreter Accuracy Results. Note that AU = Australia, AT = Austria, BR = Brazil, BG = Bulgaria, CN = China, FI = Finland, HN = Honduras, IN = India, MW = Malawi, NL = Netherlands, NZ = New Zealand, ES = Spain, SE = Sweden, US = USA, ZM = Zambia.}
\label{tab:gemini_alignment}
\end{table*}

\begin{table*}[h]
\centering
\small
\setlength{\tabcolsep}{3pt}
\renewcommand{\arraystretch}{1.05}
\begin{tabular}{l*{16}{c}}
\hline
& \multicolumn{16}{c}{\textbf{Interpreter}} \\
\cline{2-17}
\textbf{Generator}
& AU & AT & BR & BG & CN & FI & HN & IN & MW & NL & NZ & ES & SE & US & ZM & \textbf{Avg} \\
\hline
AU & \textbf{0.61} & 0.62 & 0.62 & 0.62 & 0.63 & 0.60 & 0.61 & 0.62 & 0.60 & 0.60 & 0.61 & 0.63 & 0.60 & 0.63 & 0.61 & 0.61\\
AT & 0.65 & \textbf{0.66} & 0.64 & 0.66 & 0.65 & 0.65 & 0.65 & 0.64 & 0.63 & 0.65 & 0.65 & 0.66 & 0.66 & 0.67 & 0.65 & 0.65\\
BR & 0.66 & 0.66 & \textbf{0.65} & 0.66 & 0.67 & 0.66 & 0.66 & 0.65 & 0.64 & 0.65 & 0.65 & 0.67 & 0.66 & 0.68 & 0.65 & 0.66\\
BG & 0.67 & 0.65 & 0.66 & \textbf{0.67} & 0.66 & 0.65 & 0.65 & 0.66 & 0.63 & 0.65 & 0.65 & 0.65 & 0.66 & 0.67 & 0.64 & 0.65\\
CN & 0.47 & 0.46 & 0.48 & 0.46 & \textbf{0.45} & 0.46 & 0.44 & 0.46 & 0.43 & 0.46 & 0.47 & 0.47 & 0.45 & 0.47 & 0.46 & 0.46 \\
FI & 0.60 & 0.61 & 0.62 & 0.62 & 0.61 & \textbf{0.62} & 0.61 & 0.61 & 0.60 & 0.61 & 0.60 & 0.62 & 0.62 & 0.62 & 0.61 & 0.61\\
HN & 0.70 & 0.70 & 0.70 & 0.70 & 0.70 & 0.69 & \textbf{0.71} & 0.71 & 0.68 & 0.68 & 0.68 & 0.69 & 0.69 & 0.69 & 0.69 & 0.69\\
IN & 0.73 & 0.72 & 0.73 & 0.71 & 0.73 & 0.71 & 0.72 & \textbf{0.74} & 0.71 & 0.71 & 0.71 & 0.73 & 0.71 & 0.73 & 0.72 & 0.72\\
MW & 0.75 & 0.75 & 0.75 & 0.75 & 0.75 & 0.74 & 0.76 & 0.76 & \textbf{0.75} & 0.73 & 0.74 & 0.74 & 0.74 & 0.75 & 0.75 & 0.75\\
NL & 0.61 & 0.61 & 0.61 & 0.61 & 0.61 & 0.59 & 0.60 & 0.61 & 0.60 & \textbf{0.60} & 0.60 & 0.61 & 0.59 & 0.60 & 0.60 & 0.60\\
NZ & 0.66 & 0.66 & 0.66 & 0.66 & 0.65 & 0.65 & 0.66 & 0.67 & 0.65 & 0.66 & \textbf{0.65} & 0.65 & 0.66 & 0.67 & 0.66 & 0.66\\
ES & 0.58 & 0.58 & 0.58 & 0.58 & 0.60 & 0.59 & 0.57 & 0.59 & 0.56 & 0.57 & 0.57 & \textbf{0.58} & 0.58 & 0.57 & 0.57 & 0.58\\
SE & 0.60 & 0.61 & 0.61 & 0.61 & 0.63 & 0.60 & 0.61 & 0.60 & 0.60 & 0.60 & 0.60 & 0.61 & \textbf{0.61} & 0.62 & 0.59 & 0.61\\
US & 0.62 & 0.59 & 0.62 & 0.61 & 0.63 & 0.61 & 0.60 & 0.62 & 0.59 & 0.61 & 0.60 & 0.63 & 0.61 & \textbf{0.62} & 0.61 & 0.61\\
ZM & 0.61 & 0.61 & 0.62 & 0.59 & 0.61 & 0.58 & 0.60 & 0.59 & 0.58 & 0.60 & 0.60 & 0.61 & 0.59 & 0.63 & \textbf{0.59} & 0.60\\
\hline
\textbf{Avg} & 0.63 & 0.63 & 0.64 & 0.63 & 0.64 & 0.63 & 0.63 & 0.64 & 0.62 & 0.63 & 0.63 & 0.64 & 0.63 & 0.64 & 0.63 & 0.63\\ 
\hline
\end{tabular}
\caption{Mistral Model: Generator-Interpreter Accuracy Results. Note that AU = Australia, AT = Austria, BR = Brazil, BG = Bulgaria, CN = China, FI = Finland, HN = Honduras, IN = India, MW = Malawi, NL = Netherlands, NZ = New Zealand, ES = Spain, SE = Sweden, US = USA, ZM = Zambia.}
\label{tab:mistral_alignment}
\end{table*}

\begin{table*}[h]
\centering
\small
\setlength{\tabcolsep}{3pt}
\renewcommand{\arraystretch}{1.05}
\begin{tabular}{l*{16}{c}}
\hline
& \multicolumn{16}{c}{\textbf{Interpreter}} \\
\cline{2-17}
\textbf{Generator}
& AU & AT & BR & BG & CN & FI & HN & IN & MW & NL & NZ & ES & SE & US & ZM & \textbf{Avg}\\
\hline
AU & \textbf{0.74} & 0.74 & 0.74 & 0.73 & 0.74 & 0.73 & 0.73 & 0.74 & 0.72 & 0.74 & 0.73 & 0.75 & 0.74 & 0.75 & 0.73 & 0.74\\
AT & 0.80 & \textbf{0.81} & 0.81 & 0.81 & 0.82 & 0.79 & 0.80 & 0.79 & 0.77 & 0.82 & 0.80 & 0.82 & 0.82 & 0.80 & 0.79 & 0.80\\
BR & 0.78 & 0.77 & \textbf{0.75} & 0.77 & 0.78 & 0.77 & 0.77 & 0.75 & 0.76 & 0.77 & 0.76 & 0.77 & 0.78 & 0.77 & 0.76 & 0.77\\
BG & 0.76 & 0.77 & 0.75 & \textbf{0.76} & 0.78 & 0.75 & 0.76 & 0.76 & 0.76 & 0.76 & 0.76 & 0.75 & 0.76 & 0.77 & 0.75 & 0.76\\
CN & 0.64 & 0.62 & 0.61 & 0.65 & \textbf{0.65} & 0.62 & 0.64 & 0.64 & 0.62 & 0.63 & 0.62 & 0.64 & 0.62 & 0.63 & 0.62 & 0.63\\
FI & 0.72 & 0.74 & 0.72 & 0.74 & 0.74 & \textbf{0.73} & 0.71 & 0.72 & 0.71 & 0.71 & 0.71 & 0.73 & 0.71 & 0.71 & 0.71 & 0.72\\
HN & 0.77 & 0.78 & 0.78 & 0.78 & 0.78 & 0.76 & \textbf{0.77} & 0.78 & 0.77 & 0.77 & 0.75 & 0.78 & 0.76 & 0.79 & 0.77 & 0.77\\
IN & 0.80 & 0.80 & 0.79 & 0.79 & 0.80 & 0.79 & 0.79 & \textbf{0.81} & 0.79 & 0.80 & 0.78 & 0.80 & 0.79 & 0.80 & 0.81 & 0.80\\
MW & 0.80 & 0.80 & 0.79 & 0.81 & 0.81 & 0.81 & 0.81 & 0.80 & \textbf{0.81} & 0.80 & 0.81 & 0.81 & 0.78 & 0.83 & 0.81 & 0.81\\
NL & 0.71 & 0.74 & 0.72 & 0.72 & 0.74 & 0.70 & 0.70 & 0.72 & 0.71 & \textbf{0.73} & 0.73 & 0.74 & 0.72 & 0.74 & 0.70 & 0.72\\
NZ & 0.75 & 0.76 & 0.74 & 0.74 & 0.75 & 0.74 & 0.74 & 0.76 & 0.73 & 0.73 & \textbf{0.75} & 0.74 & 0.71 & 0.76 & 0.72 & 0.74\\
ES & 0.75 & 0.77 & 0.77 & 0.78 & 0.76 & 0.75 & 0.76 & 0.75 & 0.75 & 0.76 & 0.75 & \textbf{0.76} & 0.76 & 0.76 & 0.75 & 0.76\\
SE & 0.70 & 0.72 & 0.69 & 0.71 & 0.74 & 0.70 & 0.70 & 0.70 & 0.71 & 0.71 & 0.71 & 0.72 & \textbf{0.70} & 0.70 & 0.70 & 0.71\\
US & 0.70 & 0.71 & 0.71 & 0.71 & 0.73 & 0.73 & 0.71 & 0.71 & 0.70 & 0.73 & 0.70 & 0.71 & 0.71 & \textbf{0.72} & 0.72 & 0.71\\
ZM & 0.68 & 0.69 & 0.69 & 0.70 & 0.71 & 0.67 & 0.67 & 0.67 & 0.69 & 0.68 & 0.67 & 0.70 & 0.69 & 0.69 & \textbf{0.68} & 0.69\\
\hline
\textbf{Avg} & 0.74 & 0.75 & 0.74 & 0.75 & 0.76 & 0.74 & 0.74 & 0.74 & 0.73 & 0.74 & 0.74 & 0.75 & 0.74 & 0.75 & 0.73 & 0.74\\ 
\hline
\end{tabular}
\caption{DeepSeek Model: Generator-Interpreter Accuracy Results. Note that AU = Australia, AT = Austria, BR = Brazil, BG = Bulgaria, CN = China, FI = Finland, HN = Honduras, IN = India, MW = Malawi, NL = Netherlands, NZ = New Zealand, ES = Spain, SE = Sweden, US = USA, ZM = Zambia.}
\label{tab:deepseek_alignment}
\end{table*}

\begin{table*}[t]
\centering
\small
\setlength{\tabcolsep}{3pt}
\renewcommand{\arraystretch}{1.05}
\begin{tabular}{l*{16}{c}}
\hline
& \multicolumn{16}{c}{\textbf{Interpreter}} \\
\cline{2-17}
\textbf{Generator}
& AU & AT & BR & BG & CN & FI & HN & IN & MW & NL & NZ & ES & SE & US & ZM & \textbf{Avg} \\
\hline
AU & \textbf{0.68} & 0.69 & 0.69 & 0.69 & 0.68 & 0.70 & 0.69 & 0.71 & 0.68 & 0.70 & 0.69 & 0.68 & 0.69 & 0.69 & 0.68 & 0.69\\
AT & 0.73 & \textbf{0.74} & 0.74 & 0.73 & 0.73 & 0.73 & 0.73 & 0.75 & 0.73 & 0.75 & 0.73 & 0.72 & 0.73 & 0.73 & 0.74 & 0.73\\
BR & 0.69 & 0.70 & \textbf{0.70} & 0.70 & 0.71 & 0.71 & 0.70 & 0.70 & 0.69 & 0.70 & 0.71 & 0.71 & 0.70 & 0.70 & 0.68 & 0.70\\
BG & 0.73 & 0.74 & 0.72 & \textbf{0.73} & 0.73 & 0.73 & 0.73 & 0.73 & 0.74 & 0.73 & 0.72 & 0.74 & 0.74 & 0.71 & 0.72 & 0.73\\
CN & 0.56 & 0.56 & 0.56 & 0.55 & \textbf{0.52} & 0.55 & 0.55 & 0.57 & 0.53 & 0.56 & 0.55 & 0.55 & 0.55 & 0.55 & 0.54 & 0.55\\
FI & 0.69 & 0.71 & 0.71 & 0.71 & 0.69 & \textbf{0.70} & 0.69 & 0.69 & 0.69 & 0.71 & 0.69 & 0.70 & 0.70 & 0.69 & 0.69 & 0.70\\
HN & 0.74 & 0.74 & 0.73 & 0.75 & 0.73 & 0.74 & \textbf{0.74} & 0.73 & 0.72 & 0.74 & 0.75 & 0.75 & 0.75 & 0.76 & 0.72 & 0.74\\
IN & 0.76 & 0.77 & 0.75 & 0.76 & 0.75 & 0.76 & 0.76 & \textbf{0.77} & 0.76 & 0.75 & 0.75 & 0.75 & 0.77 & 0.77 & 0.75 & 0.76\\
MW & 0.78 & 0.77 & 0.79 & 0.78 & 0.79 & 0.77 & 0.79 & 0.79 & \textbf{0.80} & 0.78 & 0.78 & 0.77 & 0.79 & 0.79 & 0.79 & 0.78\\
NL & 0.70 & 0.69 & 0.70 & 0.70 & 0.68 & 0.69 & 0.69 & 0.69 & 0.70 & \textbf{0.70} & 0.70 & 0.70 & 0.70 & 0.69 & 0.70 & 0.70\\
NZ & 0.70 & 0.69 & 0.69 & 0.70 & 0.70 & 0.69 & 0.69 & 0.68 & 0.69 & 0.68 & \textbf{0.69} & 0.68 & 0.69 & 0.68 & 0.69 & 0.69\\
ES & 0.67 & 0.68 & 0.67 & 0.67 & 0.64 & 0.65 & 0.65 & 0.65 & 0.67 & 0.66 & 0.67 & \textbf{0.67} & 0.65 & 0.66 & 0.66 & 0.66\\
SE & 0.67 & 0.67 & 0.68 & 0.68 & 0.68 & 0.69 & 0.67 & 0.67 & 0.68 & 0.67 & 0.67 & 0.69 & \textbf{0.68} & 0.65 & 0.65 & 0.67\\
US & 0.67 & 0.68 & 0.67 & 0.68 & 0.66 & 0.67 & 0.66 & 0.68 & 0.66 & 0.67 & 0.68 & 0.67 & 0.68 & \textbf{0.67} & 0.67 & 0.67\\
ZM & 0.70 & 0.71 & 0.68 & 0.69 & 0.69 & 0.69 & 0.67 & 0.69 & 0.70 & 0.71 & 0.70 & 0.68 & 0.71 & 0.69 & \textbf{0.69} & 0.69\\
\hline
\textbf{Avg} & 0.70 & 0.70 & 0.70 & 0.70 & 0.69 & 0.70 & 0.69 & 0.70 & 0.69 & 0.70 & 0.70 & 0.70 & 0.70 & 0.70 & 0.69 & 0.70\\ 
\hline
\end{tabular}
\caption{Qwen2 Model: Generator-Interpreter Accuracy Results. Note that AU = Australia, AT = Austria, BR = Brazil, BG = Bulgaria, CN = China, FI = Finland, HN = Honduras, IN = India, MW = Malawi, NL = Netherlands, NZ = New Zealand, ES = Spain, SE = Sweden, US = USA, ZM = Zambia.}
\label{tab:qwen_alignment}
\end{table*}

\begin{table*}[h]
\centering
\small
\setlength{\tabcolsep}{3pt}
\renewcommand{\arraystretch}{1.05}
\begin{tabular}{l*{16}{c}}
\hline
& \multicolumn{16}{c}{\textbf{Interpreter}} \\
\cline{2-17}
\textbf{Generator}
& AU & AT & BR & BG & CN & FI & HN & IN & MW & NL & NZ & ES & SE & US & ZM & \textbf{Avg}\\
\hline
AU & \textbf{0.76} & 0.77 & 0.73 & 0.75 & 0.73 & 0.71 & 0.72 & 0.71 & 0.69 & 0.75 & 0.74 & 0.74 & 0.71 & 0.75 & 0.70 & 0.73 \\
AT & 0.77 & \textbf{0.77} & 0.77 & 0.77 & 0.73 & 0.72 & 0.74 & 0.75 & 0.69 & 0.77 & 0.77 & 0.77 & 0.75 & 0.79 & 0.71 & 0.75\\
BR & 0.70 & 0.67 & \textbf{0.69} & 0.69 & 0.66 & 0.66 & 0.68 & 0.67 & 0.64 & 0.68 & 0.69 & 0.68 & 0.67 & 0.71 & 0.66 & 0.68\\
BG & 0.73 & 0.71 & 0.70 & \textbf{0.73} & 0.71 & 0.69 & 0.69 & 0.71 & 0.68 & 0.71 & 0.72 & 0.71 & 0.69 & 0.72 & 0.68 & 0.71\\
CN & 0.59 & 0.59 & 0.59 & 0.59 & \textbf{0.61} & 0.57 & 0.58 & 0.59 & 0.57 & 0.58 & 0.59 & 0.59 & 0.55 & 0.61 & 0.57 & 0.58\\
FI & 0.72 & 0.73 & 0.70 & 0.73 & 0.71 & \textbf{0.70} & 0.68 & 0.70 & 0.68 & 0.71 & 0.73 & 0.71 & 0.70 & 0.75 & 0.69 & 0.71\\
HN & 0.71 & 0.71 & 0.72 & 0.71 & 0.68 & 0.65 & \textbf{0.71} & 0.68 & 0.69 & 0.69 & 0.71 & 0.71 & 0.67 & 0.73 & 0.69 & 0.70\\
IN & 0.71 & 0.69 & 0.70 & 0.71 & 0.66 & 0.67 & 0.68 & \textbf{0.72} & 0.69 & 0.70 & 0.71 & 0.69 & 0.69 & 0.70 & 0.69 & 0.69\\
MW & 0.67 & 0.65 & 0.65 & 0.63 & 0.62 & 0.62 & 0.65 & 0.65 & \textbf{0.63} & 0.64 & 0.67 & 0.65 & 0.61 & 0.67 & 0.63 & 0.64\\
NL & 0.75 & 0.76 & 0.72 & 0.74 & 0.71 & 0.72 & 0.73 & 0.72 & 0.71 & \textbf{0.74} & 0.73 & 0.73 & 0.72 & 0.74 & 0.70 & 0.73\\
NZ & 0.69 & 0.66 & 0.66 & 0.66 & 0.64 & 0.64 & 0.66 & 0.67 & 0.63 & 0.66 & \textbf{0.68} & 0.65 & 0.65 & 0.69 & 0.64 & 0.66\\
ES & 0.73 & 0.72 & 0.73 & 0.73 & 0.69 & 0.67 & 0.70 & 0.71 & 0.66 & 0.72 & 0.72 & \textbf{0.72} & 0.69 & 0.75 & 0.68 & 0.71\\
SE & 0.69 & 0.70 & 0.69 & 0.73 & 0.68 & 0.66 & 0.69 & 0.69 & 0.67 & 0.66 & 0.66 & 0.67 & \textbf{0.67} & 0.71 & 0.69 & 0.68\\
US & 0.70 & 0.68 & 0.68 & 0.69 & 0.67 & 0.64 & 0.69 & 0.67 & 0.67 & 0.67 & 0.68 & 0.70 & 0.64 & \textbf{0.71} & 0.68 & 0.68\\
ZM & 0.66 & 0.67 & 0.67 & 0.66 & 0.67 & 0.63 & 0.67 & 0.67 & 0.66 & 0.65 & 0.66 & 0.65 & 0.62 & 0.65 & \textbf{0.66} & 0.66\\
\hline
\textbf{Avg} & 0.71 & 0.70 & 0.69 & 0.70 & 0.68 & 0.66 & 0.68 & 0.69 & 0.66 & 0.69 & 0.70 & 0.69 & 0.67 & 0.71 & 0.67 & 0.69\\ 
\hline
\end{tabular}
\caption{Claude Model: Generator-Interpreter Accuracy Results. Note that AU = Australia, AT = Austria, BR = Brazil, BG = Bulgaria, CN = China, FI = Finland, HN = Honduras, IN = India, MW = Malawi, NL = Netherlands, NZ = New Zealand, ES = Spain, SE = Sweden, US = USA, ZM = Zambia.}
\label{tab:claude_alignment}

\end{table*}

\section{Confusion Matrices for Generator Countries}\label{sec:app_diagnostics}

As discussed in Section~\ref{sec:generator-interpreter-alignment}, some generator countries consistently yield higher interpretation accuracy, while others present substantial challenges across models. To provide a more balanced view, we present confusion matrices for four representative countries: the top two generators with highest interpretation accuracy (Austria, Malawi) and the two generators with lowest accuracy (China, Sweden). For each generator country, we include confusion matrices for the top-3 best-performing interpreter personas and the bottom-3 worst-performing interpreter personas in the GPT-4o model.

\subsection{Austria}
\begin{figure}[h]
    \centering
    \includegraphics[width=\linewidth]{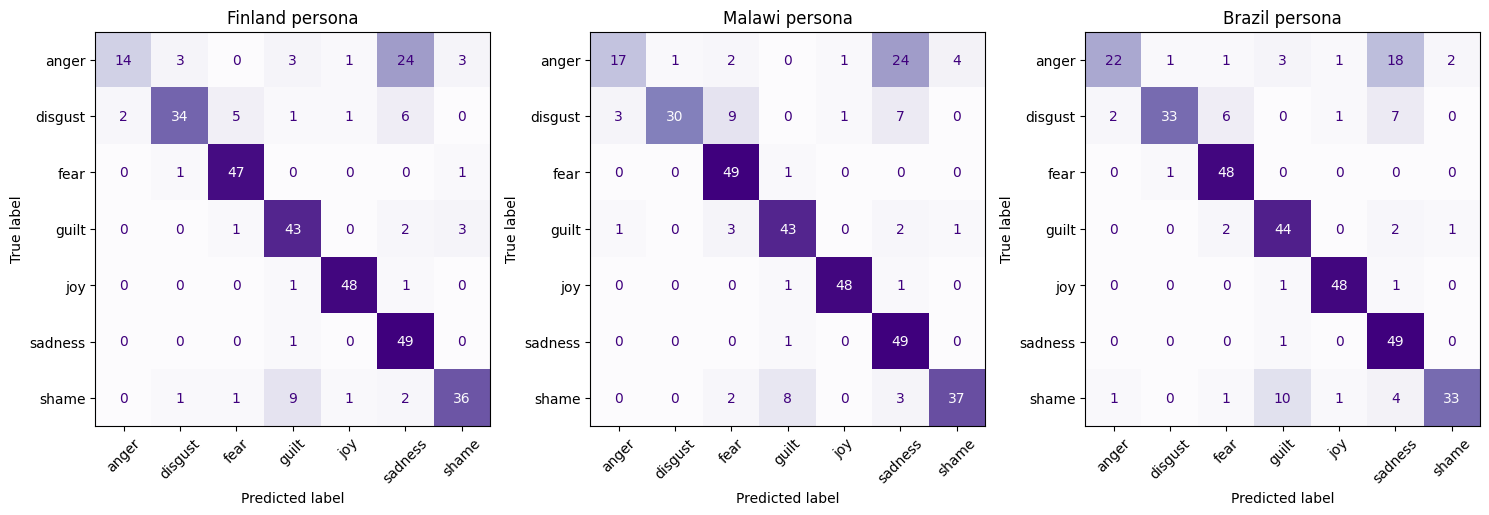}
    \caption{Worst-performing personas in the GPT-4o model (Generator: Austria).}
    \label{fig:gpt-at-worst}
\end{figure}

\begin{figure}[h]
    \centering
    \includegraphics[width=\linewidth]{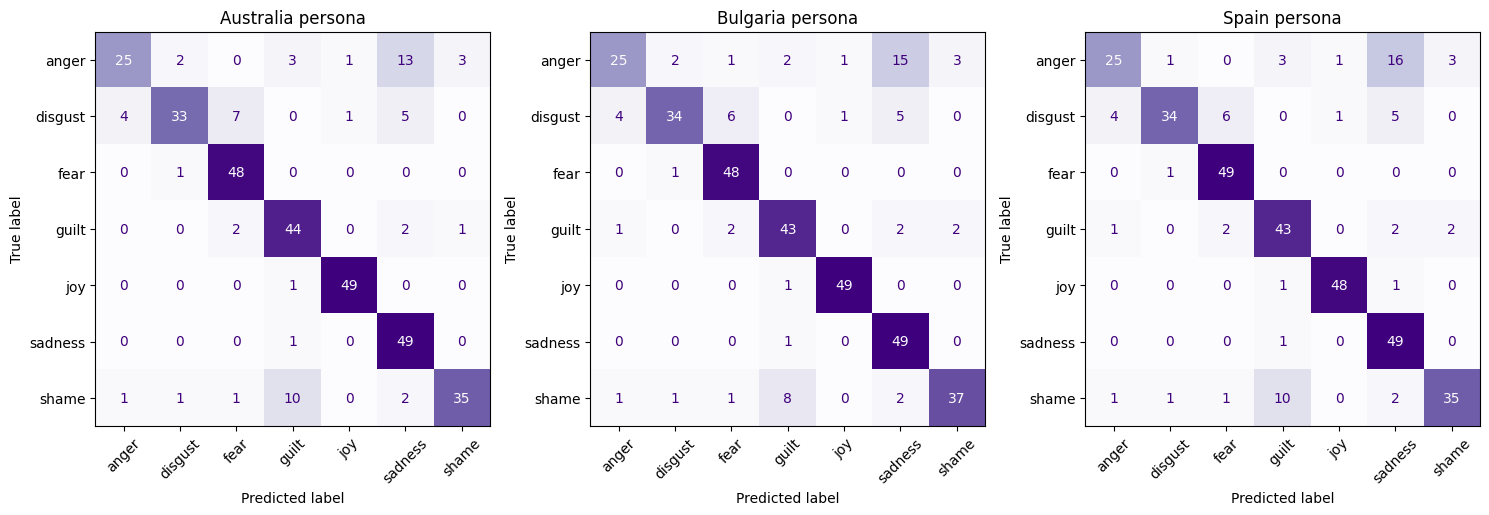}
    \caption{Best-performing personas in the GPT-4o model (Generator: Austria)}
    \label{fig:gpt-at-best}
\end{figure}

Figures~\ref{fig:gpt-at-worst} and~\ref{fig:gpt-at-best} show confusion matrices for Austrian-originated events.
The worst interpreters are Finland (0.77), Malawi (0.78), Brazil (0.79), while the best interpreters include Australia (0.81), Bulgaria (0.81), Spain (0.81).

\subsection{Malawi}
\begin{figure}[h]
    \centering
    \includegraphics[width=\linewidth]{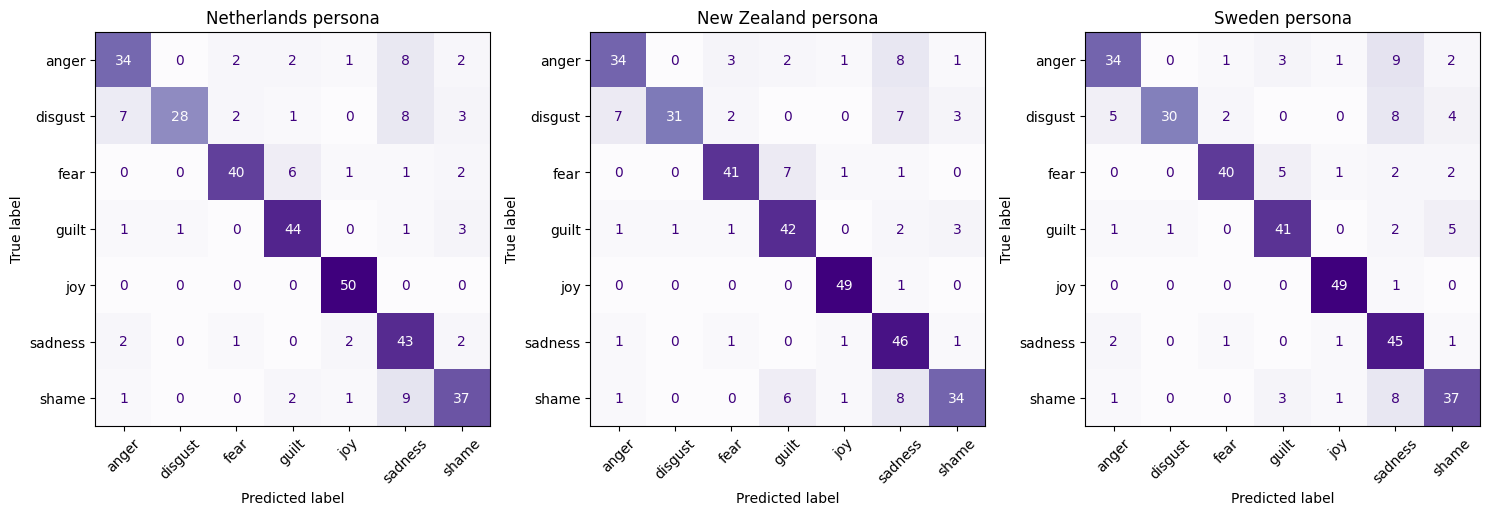}
    \caption{Worst-performing personas in the GPT-4o model (Generator: Malawi)}
    \label{fig:gpt-mw-worst}
\end{figure}

\begin{figure}[h]
    \centering
    \includegraphics[width=\linewidth]{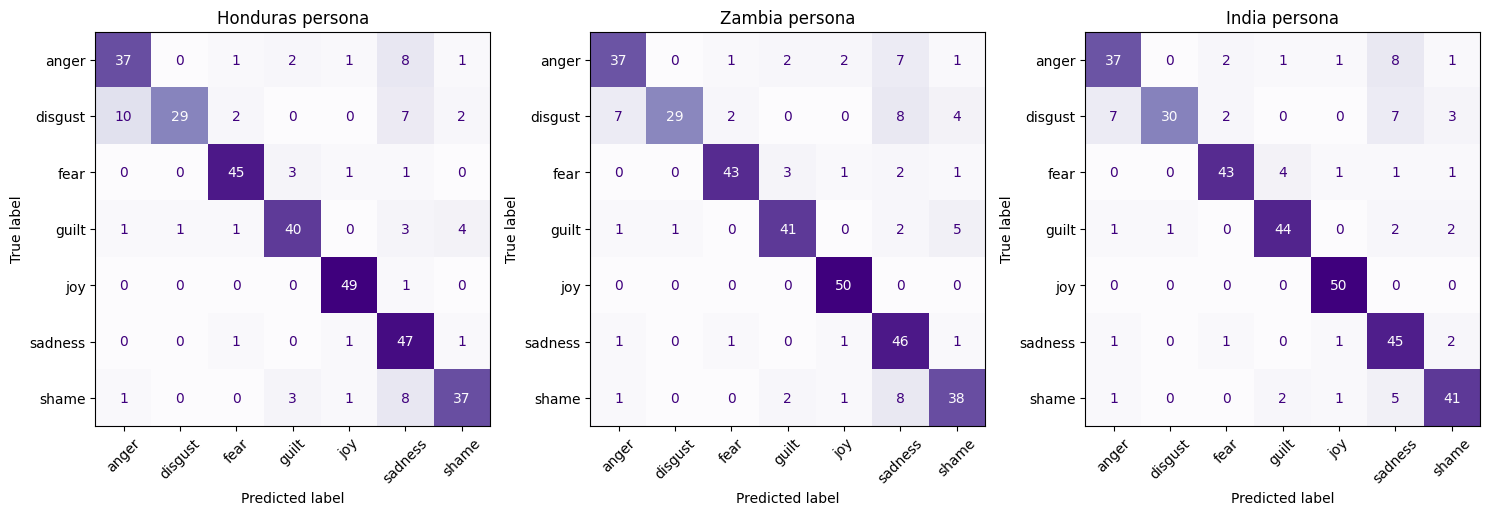}
    \caption{Best-performing personas in the GPT-4o model (Generator: Malawi)}
    \label{fig:gpt-mw-best}
\end{figure}

Figures~\ref{fig:gpt-mw-worst} and~\ref{fig:gpt-mw-best} illustrate Malawi-originated events. The worst interpreters are the Netherlands (0.79), New Zealand (0.79), Sweden (0.79), while the best interpreters are Honduras (0.81), Zambia (0.81), India (0.83).  

\subsection{China}
\begin{figure}[h]
    \centering
    \includegraphics[width=\linewidth]{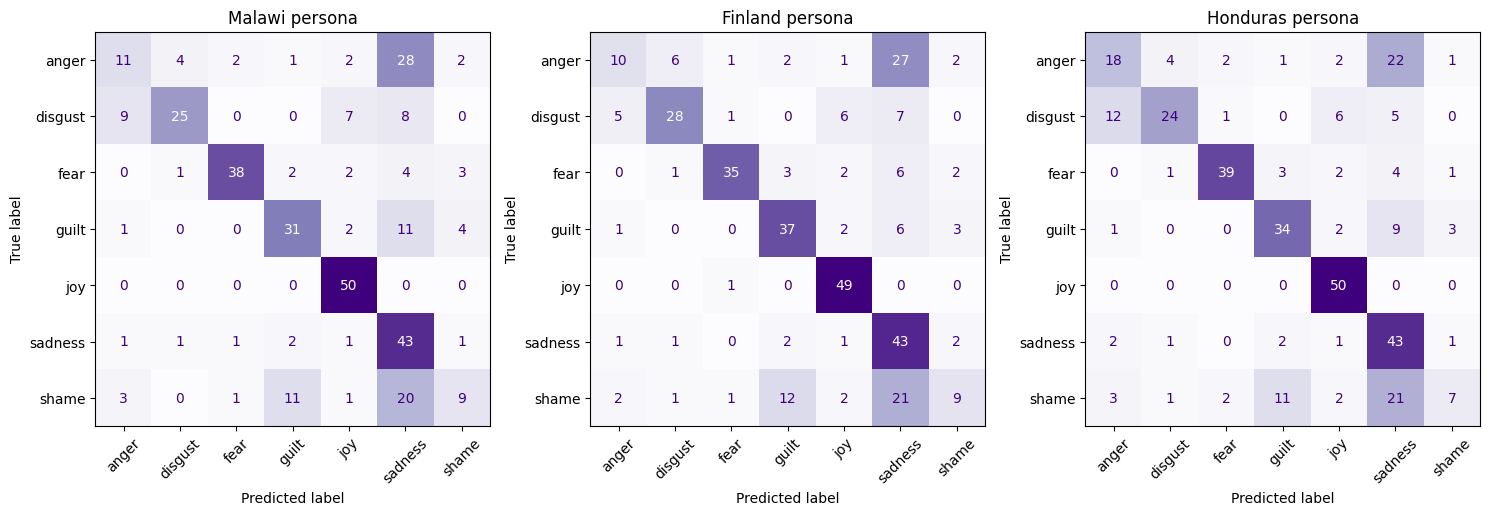}
    \caption{Worst-performing personas in the GPT-4o model (Generator: China)}
    \label{fig:gpt-cn-worst}
\end{figure}

\begin{figure}[h]
    \centering
    \includegraphics[width=\linewidth]{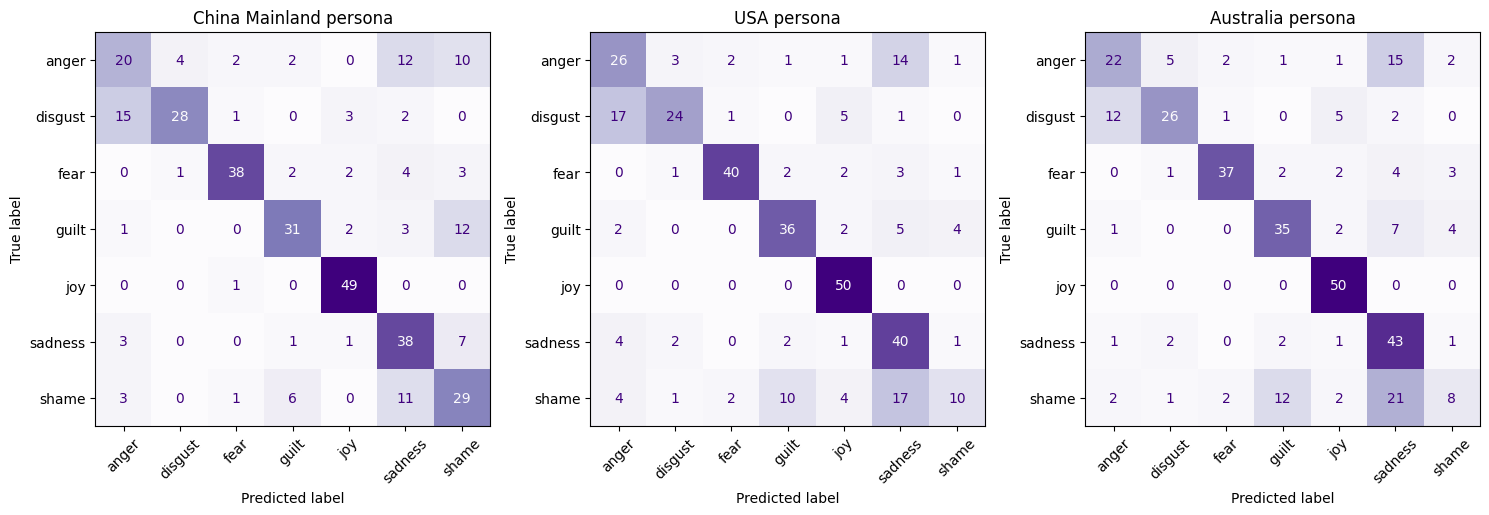}
    \caption{Best-performing personas in the GPT-4o model (Generator: China)}
    \label{fig:gpt-cn-best}
\end{figure}

Figures~\ref{fig:gpt-cn-worst} and~\ref{fig:gpt-cn-best} illustrate Chinese-originated events. The worst interpreters are Malawi (0.59), Finland (0.60), Honduras (0.61), while the best interpreters are  Australia (0.63), USA (0.65), China (0.67).

\subsection{Sweden}
\begin{figure}[h]
    \centering
    \includegraphics[width=\linewidth]{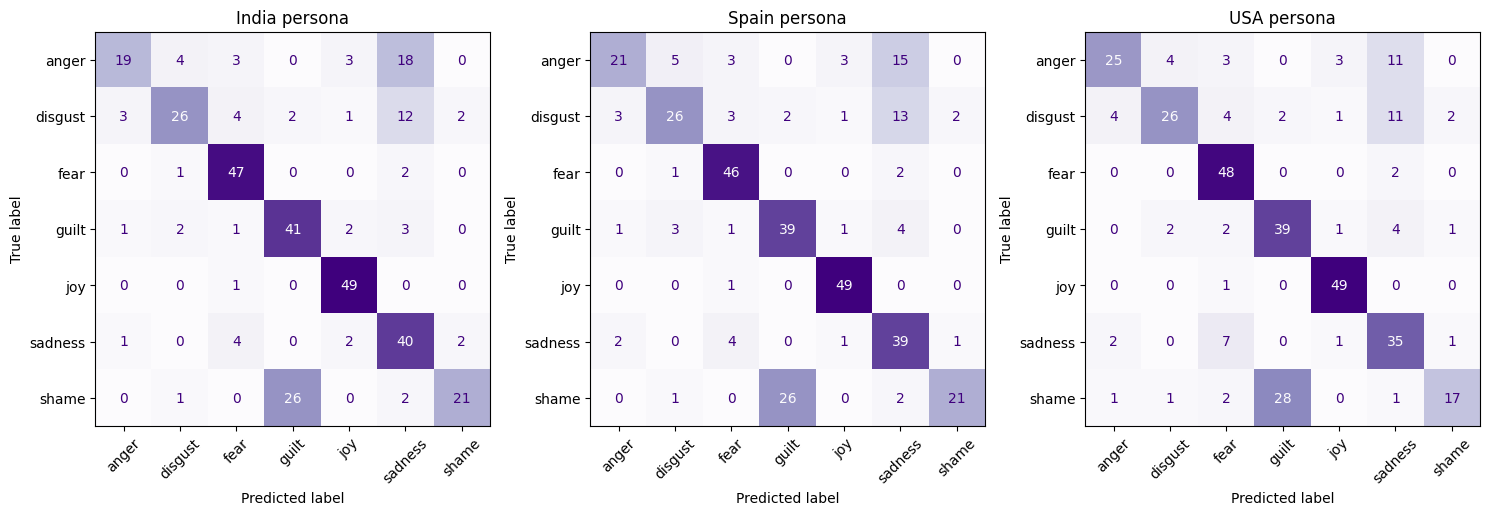}
    \caption{Worst-performing personas in the GPT-4o model (Generator: Sweden)}
    \label{fig:gpt-se-worst}
\end{figure}

\begin{figure}[h]
    \centering
    \includegraphics[width=\linewidth]{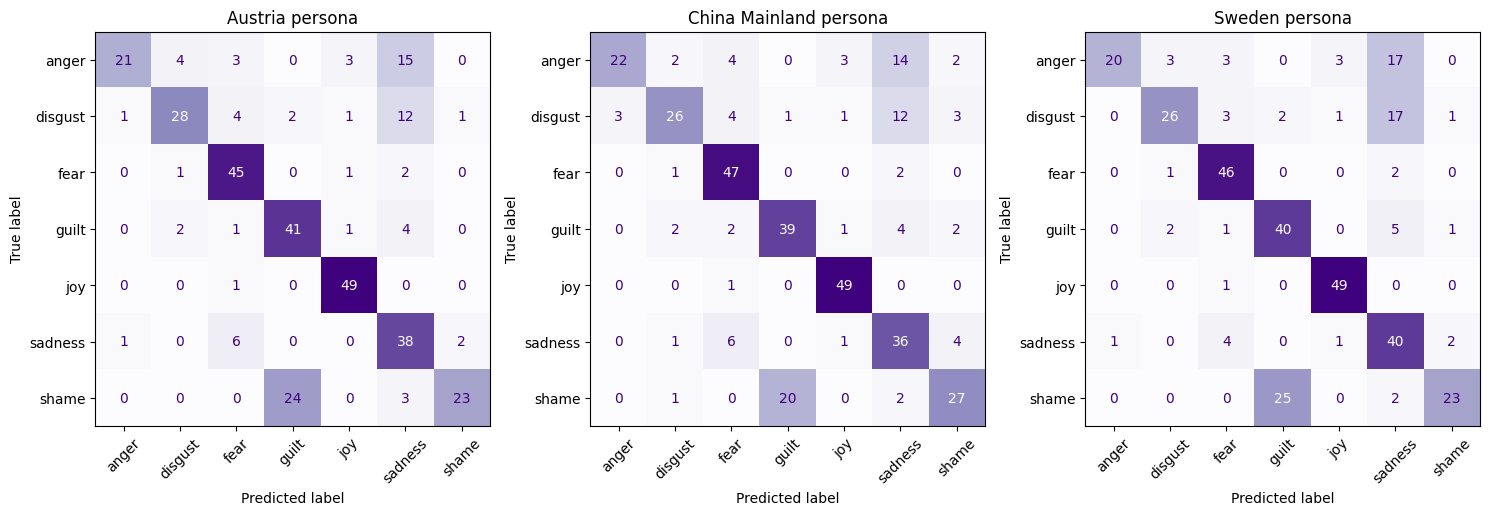}
    \caption{Best-performing personas in the GPT-4o model (Generator: Sweden)}
    \label{fig:gpt-se-best}
\end{figure}

Figures~\ref{fig:gpt-se-worst} and~\ref{fig:gpt-se-best} illustrate Sweden-originated events. The worst interpreters are USA (0.68), India (0.69), Spain (0.69), while the best interpreters are  Austria (0.70), China (0.70), Sweden (0.70).   

\subsection{Illustrative Patterns in Emotion Misattribution for Chinese-Origin Events}

%Table~\ref{tab:gpt-cn-predictions} provides illustrative examples of Chinese-origin events with gold-label emotions alongside predictions from multiple personas, highlighting how interpretations diverge across cultural contexts.

Table~\ref{tab:gpt-cn-predictions} provides illustrative examples showing that emotion misattributions for Chinese-origin events are systematic and culturally patterned rather than random.

First, self-conscious emotions frequently shift between shame and guilt across cultural interpretations. For example, in ``Failing in an exam because of laziness,'' the gold-label emotion is shame, reflecting concern over social evaluation and loss of face. However, most non-Chinese personas reinterpret this event as guilt, framing it as an individual moral failure. This pattern suggests a transition from socially oriented self-evaluation to individual responsibility-based reasoning.

Second, anger is consistently attenuated into inward-oriented emotions. In morally charged or victimization scenarios such as ``Being sexually assaulted on a bus and no one helped,'' the gold-label anger, emphasizing injustice and social failure, is replaced by fear, sadness, or shame across personas. Similarly, in ``Being insulted in front of my parents,'' anger is reinterpreted as sadness or shame, indicating a tendency to internalize relational violations rather than recognize outward-directed moral anger.

Third, collective and relational meanings are often lost in cross-cultural interpretation. In the example ``South Korea winning more gold medals than China,'' the gold-label emotion is anger, reflecting relative national status concerns. Yet all interpreter personas predict joy, treating the event as a neutral or positive sports outcome and overlooking its competitive relational framing.

Finally, cultural and moral disgust is systematically misrecognized. In ``Saw some girls dressed up like Americans,'' disgust expresses cultural or normative disapproval, but all personas predict joy, suggesting that interpreter reasoning associates disgust primarily with physical or visceral cues rather than culturally embedded moral aversion.

Taken together, these examples show a recurring collapse of socially embedded emotions into a small set of internally oriented affective categories, supporting the broader claim that emotion attribution in LLMs is strongly shaped by the cultural context of emotion generation.

\begin{table*}[t]
\centering
\small
\renewcommand{\arraystretch}{1.2}
\begin{tabular}{p{5cm}cccccc}
\hline
\textbf{ISEAR Event Examples} & \textbf{Gold-label} & \textbf{US} & \textbf{AU} & \textbf{BG} & \textbf{NL} & \textbf{IN} \\
\hline
Failing in an exam because of laziness
& shame & guilt & guilt & shame & guilt & guilt \\
\hline
South Korea winning more gold medals in the Asian Olympics (more than China) 
& anger & joy & joy & joy & joy & joy \\
\hline
Being sexually assaulted on a bus and no one helped (people only laughed)
& anger & fear & fear & sadness & fear & shame \\
\hline
Being insulted in front of my parents, and they believed other people (those insulting me)  
& anger & sadness & sadness & shame & sadness & shame \\
\hline
Not being able to do well to be chosen for the English lecture 
& sadness & sadness & sadness & shame & sadness & shame \\
\hline
Saw some girls dressed up like Americans 
& disgust & joy & joy & joy & joy & joy \\
\hline
Went to a lecture about Chinese history and heard about the Opium War
& shame & sadness & sadness & sadness & sadness & sadness \\
\hline
Found out that my grandmother might have been misdiagnosed as a  psychiatric patient because the doctor did not know about neurologism 
& sadness & anger & sadness & sadness & sadness & sadness\\  
\hline
Said something I should not have said in public
& guilt & shame & shame & shame & shame & shame\\  
\hline
People having a football match outside the window during the lecture
& anger & anger & disgust & disgust & disgust & disgust \\  
\hline
I kicked the chair in the classroom to express my anger
& guilt & anger & anger & anger & anger & anger\\  
\hline
\end{tabular}
\caption{Examples of emotion predictions for Chinese-origin events in the GPT-4o model. The \textbf{gold-label} column shows the expected emotion, while the remaining columns display predictions from different personas. Note that US = USA, AU = Australia, BG = Bulgaria, NL = Netherlands, IN = India.}
\label{tab:gpt-cn-predictions}
\end{table*}

\section{Structured Output}\label{sec:app_structured}
Despite being explicitly prompted to select from seven predefined emotion categories, all models frequently produced outputs outside the target label set (see Table \ref{tab:unique-emotions}). For certain nationality personas, models also generated emotion labels in the corresponding local languages rather than English. Adherence to prompt constraints varied substantially across models: GPT and Qwen2 most consistently followed the categorical instruction, whereas other models frequently expanded beyond the requested output by adding unsolicited reasoning or extraneous content. Claude, in particular, exhibited higher refusal rates and produced more hallucinatory output.

To further explore output variability, we evaluated a \emph{structured output} setting in which models were required to produce responses conforming to a predefined schema.
% specifically by selecting exactly one emotion label from the seven-category set and returning it without additional text.
Structured output substantially reduced category proliferation and eliminated invalid labels, yielding outputs that strictly adhered to the seven target categories with consistent formatting and no extraneous content. While this constraint did not lead to a notable increase in overall accuracy, it consistently resulted in higher precision and recall across models.

\begin{table*}[h]
\centering
\small
\begin{tabular}{p{3.5cm} p{10.5cm}}
\hline
\textbf{Model} & \textbf{Emotions} \\
\hline
gpt-4o-2024-05-13 &
anguish, annoyance, anxiety, awkwardness, betrayal, concern, confusion, curiosity, disappointment, discomfort, disgust, embarrassment, frustration, grief, hate, indifference, irritation, regret, relief, shock, stress, surprise, worry\\
\hline
gemini-pro-1.5 &
annoyance, anticipation, anxiety, awkwardness,
boredom, concern, confusion, curiosity, disappointment, discomfort, doubt, embarrassment, envy, exasperation, frustration, gratitude, grief, hope, horror, humiliation, impatience, indifference, indignation, injustice, interest, intrigue, irritation, 
jealousy, nervousness, neutral, pain, panic, pressure, pride, regret, relief, resignation, respect, shock, stress, surprise, terror, worry
\\
\hline
claude-3.5-sonnet &
acceptance, amusement, annoyance, anticipation, anxiety, boredom, betrayal, concern, confusion, curiosity, disappointment, discomfort, embarrassment, envy, gratitude, grief, hope, humiliation, frustration, indifference, irritation, jealousy, loneliness, mix, neutral,  nostalgia, pain, peace, pride, regret, relief, respect, satisfaction, skepticism, stress, surprise, worry, unease\\
\hline
mistral-7b-instruct-v0.3 &
affection, amusement, annoyance, anticipation, anxiety, apprehension, astonishment, attentiveness, awe, boredom, calmness, caution,
compassion, concern, confusion, contentment, curiosity, determination, disappointment, discomfort, disinterest, disagreeableness, embarrassment,  empathy, excitement, exhaustion,
frustration, focus, gratitude, grief, hate, humility, indifference, interest, jealousy, melancholy, nervousness, neutral, patience, peace, pride, responsibility, relaxation, relief, respectful awe, respectful anxiety, shock, shyness, surprise, sympathy, uncertainty, unease \\
\hline
qwen-2.5-7b-instruct &
anxiety, annoyance, attention, boredom,
compassion, concern, confusion, contentment, curiosity, disappointment, discomfort, distraction, embarrassment, envy, excitement, focus, frustration, 
gratitude, grief, hate, horror, jealousy, neutral, nervousness, nostalgia, pain, patience, peace,  pride, regret,  relaxation, relief,  shock, stress, surprise, uncertainty \\
\hline
deepseek-v3   & annoyance, betrayal, boredom, calmness, compassion, concern, confusion, contentment, curiosity, disappointment, discomfort, envy, frustration, gratitude, grief, indifference, irritation, jealousy, neutral, patience, relief, stress, surprise \\
\hline
\end{tabular}
\caption{Unique emotion labels generated by large language models}
\label{tab:unique-emotions}
\end{table*}

\section{Evaluation of Explicit Generator Specification}\label{sec:app_generator_spec}
% (differences remained within 1-2\% relative to unstructured prompts), 
We examined whether explicitly informing the model about the generator’s country affected emotion attribution performance. In this condition, the interpreter model was explicitly told that the emotional event had been generated by an individual from a specific country and was asked to assign an emotion label given this background information. Across all nationality personas, performance differences were small and inconsistent (see Table~\ref{tab:generator-context}, indicating that explicit disclosure of the generator’s cultural background alone does not materially improve emotion attribution.

\begin{table*}[t]
\centering
\small
\setlength{\tabcolsep}{3pt}
\renewcommand{\arraystretch}{1.05}
\begin{tabular}{l*{16}{c}}
\hline
& \multicolumn{16}{c}{\textbf{Interpreter}} \\
\cline{2-17}
\textbf{Generator}
& AU & AT & BR & BG & CN & FI & HN & IN & MW & NL & NZ & ES & SE & US & ZM & \textbf{Avg}\\
\hline
AU (Before) & \textbf{0.74} & 0.74 & 0.74 & 0.73 & 0.74 & 0.73 & 0.73 & 0.74 & 0.72 & 0.74 & 0.73 & 0.75 & 0.74 & 0.75 & 0.73 & 0.74\\
AU (After) & 0.72 & 0.73 & 0.73 & 0.74 & 0.73 & 0.73 & 0.74 & 0.73 & 0.73 & 0.74 & 0.73 & 0.73 & 0.73 & 0.74 & 0.72 & 0.73\\
\textbf{Effect} & \textcolor{red}{-0.02} & \textcolor{red}{-0.01} & \textcolor{red}{-0.01} & \textcolor{blue}{+0.01} & \textcolor{red}{-0.01} & \textcolor{blue}{0.00} & \textcolor{blue}{+0.01} & \textcolor{red}{-0.01} & \textcolor{blue}{+0.01} & \textcolor{blue}{0.00} & \textcolor{blue}{0.00} & \textcolor{red}{-0.02} & \textcolor{red}{-0.01} & \textcolor{red}{-0.01} & \textcolor{red}{-0.01} & \textcolor{red}{-0.01}\\

\hline

AT (Before) & 0.80 & \textbf{0.81} & 0.81 & 0.81 & 0.82 & 0.79 & 0.80 & 0.79 & 0.77 & 0.82 & 0.80 & 0.82 & 0.82 & 0.80 & 0.79 & 0.80\\
AT (After) & 0.62 & 0.82 & 0.64 & 0.81 & 0.83 & 0.81 & 0.82 & 0.80 & 0.81 & 0.79 & 0.82 & 0.80 & 0.83 & 0.81 & 0.82 & 0.79\\
\textbf{Effect} & \textcolor{red}{-0.18} & \textcolor{blue}{+0.01} & \textcolor{red}{-0.17} & \textcolor{blue} {0.00} & \textcolor{blue}{+0.01} & \textcolor{blue}{+0.02} & \textcolor{blue}{+0.02} & \textcolor{blue}{+0.01} & \textcolor{blue}{+0.04} & \textcolor{red}{-0.03} & \textcolor{blue}{+0.02} & \textcolor{red}{-0.02} & \textcolor{blue}{+0.01} & \textcolor{blue}{+0.01} & \textcolor{blue}{+0.03} & \textcolor{red}{-0.01}\\

\hline
BR (Before) & 0.78 & 0.77 & \textbf{0.75} & 0.77 & 0.78 & 0.77 & 0.77 & 0.75 & 0.76 & 0.77 & 0.76 & 0.77 & 0.78 & 0.77 & 0.76 & 0.77\\

BR (After) & 0.76 & 0.77 & 0.77 & 0.78 & 0.77 & 0.77 & 0.78 & 0.77 & 0.78 & 0.77 & 0.78 & 0.78 & 0.77 & 0.78 & 0.76 & 0.77\\

\textbf{Effect} & \textcolor{red}{-0.02} & \textcolor{blue}{0.00} & \textcolor{blue} {+0.02} & \textcolor{blue}{+0.01} & \textcolor{red}{-0.01} & \textcolor{blue}{0.00} & \textcolor{blue}{+0.01} & \textcolor{blue}{+0.02} & \textcolor{blue}{+0.02} & \textcolor{blue}{0.00} & \textcolor{blue}{+0.02} & \textcolor{blue}{+0.01} & \textcolor{red}{-0.01} & \textcolor{blue}{+0.01} & \textcolor{blue}{0.00} & \textcolor{blue}{0.00}\\

\hline
BG (Before) & 0.76 & 0.77 & 0.75 & \textbf{0.76} & 0.78 & 0.75 & 0.76 & 0.76 & 0.76 & 0.76 & 0.76 & 0.75 & 0.76 & 0.77 & 0.75 & 0.76\\

BG (After) & 0.76 & 0.76 & 0.76 & 0.77 & 0.77 & 0.76 & 0.76 & 0.76 & 0.74 & 0.76 & 0.75 & 0.76 & 0.76 & 0.75 & 0.75 & 0.76\\

\textbf{Effect} & \textcolor{blue}{0.00} & \textcolor{red}{-0.01} & \textcolor{blue}{+0.01} & \textcolor{blue}{+0.01} & \textcolor{red}{-0.01} & \textcolor{blue}{+0.01} & \textcolor{blue}{0.00} & \textcolor{blue}{0.00} & \textcolor{red}{-0.02} & \textcolor{blue}{0.00} & \textcolor{red}{-0.01} & \textcolor{blue}{+0.01} & \textcolor{blue}{0.00} & \textcolor{red}{-0.02} & \textcolor{blue}{0.00} & \textcolor{blue}{0.00}\\

\hline
CN (Before) & 0.64 & 0.62 & 0.61 & 0.65 & \textbf{0.65} & 0.62 & 0.64 & 0.64 & 0.62 & 0.63 & 0.62 & 0.64 & 0.62 & 0.63 & 0.62 & 0.63\\

CN (After) & 0.62 & 0.64 & 0.64 & 0.63 & 0.64 & 0.60 & 0.62 & 0.65 & 0.62 & 0.61 & 0.64 & 0.64 & 0.63 & 0.63 & 0.63 & 0.63\\

\textbf{Effect} & \textcolor{red}{-0.02} & \textcolor{blue}{+0.02} & \textcolor{blue} {+0.03} & \textcolor{red}{-0.02} & \textcolor{red}{-0.01} & \textcolor{red}{-0.02} & \textcolor{red}{-0.02} & \textcolor{blue}{+0.01} & \textcolor{blue}{0.00} & \textcolor{red}{-0.02} & \textcolor{blue}{+0.02} & \textcolor{blue}{0.00} & \textcolor{blue}{+0.01} & \textcolor{blue}{0.00} & \textcolor{blue}{+0.01} & \textcolor{blue}{0.00}\\

\hline
FI (Before) & 0.72 & 0.74 & 0.72 & 0.74 & 0.74 & \textbf{0.73} & 0.71 & 0.72 & 0.71 & 0.71 & 0.71 & 0.73 & 0.71 & 0.71 & 0.71 & 0.72\\
FI (After) & 0.72 & 0.73 & 0.72 & 0.73 & 0.72 & 0.70 & 0.72 & 0.71 & 0.71 & 0.72 & 0.71 & 0.74 & 0.71 & 0.72 & 0.73 & 0.72  \\

\textbf{Effect} & \textcolor{blue}{0.00} & \textcolor{red}{-0.01} & \textcolor{blue}{0.00}  & \textcolor{red}{-0.01}  & \textcolor{red}{-0.02}  & \textcolor{red}{-0.03} & \textcolor{blue}{+0.01} & \textcolor{red}{-0.01} & \textcolor{blue}{0.00} & \textcolor{blue}{+0.01} & \textcolor{blue}{0.00} & \textcolor{blue}{+0.01} & \textcolor{blue}{0.00} & \textcolor{blue}{+0.01} & \textcolor{blue}{+0.02} & \textcolor{blue}{0.00}\\

\hline
HN (Before) & 0.77 & 0.78 & 0.78 & 0.78 & 0.78 & 0.76 & \textbf{0.77} & 0.78 & 0.77 & 0.77 & 0.75 & 0.78 & 0.76 & 0.79 & 0.77 & 0.77\\
HN (After) & 0.76 & 0.77 & 0.77 & 0.77 & 0.77 & 0.76 & 0.78 & 0.77 & 0.76 & 0.77 & 0.75 & 0.79 & 0.77 & 0.77 & 0.77 & 0.77\\

\textbf{Effect} & \textcolor{red}{-0.01} & \textcolor{red}{-0.01} & \textcolor{red}{-0.01}  & \textcolor{red}{-0.01}  & \textcolor{red}{-0.01}  & \textcolor{blue}{0.00} & \textcolor{blue}{+0.01} & \textcolor{red}{-0.01} & \textcolor{red}{-0.01} & \textcolor{blue}{0.00} & \textcolor{blue}{0.00} & \textcolor{blue}{+0.01} & \textcolor{blue}{+0.01} & \textcolor{red}{-0.02} & \textcolor{blue}{0.00} & \textcolor{blue}{0.00}\\

\hline
IN (Before) & 0.80 & 0.80 & 0.79 & 0.79 & 0.80 & 0.79 & 0.79 & \textbf{0.81} & 0.79 & 0.80 & 0.78 & 0.80 & 0.79 & 0.80 & 0.81 & 0.80\\

IN (After) & 0.79 & 0.80 & 0.78 & 0.81 & 0.79 & 0.78 & 0.79 & 0.80 & 0.79 & 0.79 & 0.80 & 0.80 & 0.79 & 0.79 & 0.79 & 0.79 \\

\textbf{Effect} & \textcolor{red}{-0.01} & \textcolor{blue}{0.00} & \textcolor{red}{-0.01}  & \textcolor{blue}{+0.02}  & \textcolor{red}{-0.01}  & \textcolor{red}{-0.01} & \textcolor{blue}{0.00} & \textcolor{red}{-0.01} & \textcolor{blue}{0.00} & \textcolor{red}{-0.01} & \textcolor{blue}{+0.02} & \textcolor{blue}{0.00} & \textcolor{blue}{0.00} & \textcolor{red}{-0.01} & \textcolor{red}{-0.02} & \textcolor{red}{-0.01}\\

\hline
MW (Before) & 0.80 & 0.80 & 0.79 & 0.81 & 0.81 & 0.81 & 0.81 & 0.80 & \textbf{0.81} & 0.80 & 0.81 & 0.81 & 0.78 & 0.83 & 0.81 & 0.81\\

MW (After) & 0.80 & 0.81 & 0.80  & 0.81  & 0.80 & 0.80 & 0.81 & 0.80  & 0.81  & 0.81  & 0.79  & 0.80  & 0.80  & 0.81  & 0.81 & 0.80 \\

\textbf{Effect} & \textcolor{blue}{0.00} & \textcolor{blue}{+0.01} & \textcolor{blue}{+0.01} & \textcolor{blue} {0.00} & \textcolor{red}{-0.01} & \textcolor{red}{-0.01} & \textcolor{blue}{0.00} & \textcolor{blue}{0.00} & \textcolor{blue}{0.00} & \textcolor{blue}{+0.01} & \textcolor{red}{-0.02} & \textcolor{red}{-0.01} & \textcolor{blue}{+0.02} & \textcolor{red}{-0.02} & \textcolor{blue}{0.00} & \textcolor{red}{-0.01}\\

\hline
NL (Before) & 0.71 & 0.74 & 0.72 & 0.72 & 0.74 & 0.70 & 0.70 & 0.72 & 0.71 & \textbf{0.73} & 0.73 & 0.74 & 0.72 & 0.74 & 0.70 & 0.72\\
NL (After) & 0.71 & 0.73 & 0.71 & 0.73 & 0.71 & 0.70 & 0.72 & 0.71 & 0.71 & 0.72 & 0.70 & 0.72 & 0.71 & 0.71 & 0.74 & 0.72\\

\textbf{Effect} & \textcolor{blue}{0.00} & \textcolor{red}{-0.01} & \textcolor{red}{-0.01} & \textcolor{blue}{+0.01} & \textcolor{red}{-0.03} & \textcolor{blue}{0.00} & \textcolor{blue}{+0.02} & \textcolor{red}{-0.01} & \textcolor{blue}{0.00} & \textcolor{red}{-0.01} & \textcolor{red}{-0.03} & \textcolor{red}{-0.02} & \textcolor{red}{-0.01} & \textcolor{red}{-0.03} & \textcolor{blue}{+0.04} & \textcolor{blue}{0.00}\\

\hline

NZ (Before) & 0.75 & 0.76 & 0.74 & 0.74 & 0.75 & 0.74 & 0.74 & 0.76 & 0.73 & 0.73 & \textbf{0.75} & 0.74 & 0.71 & 0.76 & 0.72 & 0.74\\
NZ (After) & 0.74 & 0.74 & 0.72 & 0.74 & 0.73 & 0.71 & 0.74 & 0.73 & 0.73 & 0.74 & 0.74 & 0.74 & 0.72 & 0.74 & 0.72 & 0.73 \\

\textbf{Effect} & \textcolor{red}{-0.01} & \textcolor{red}{-0.02} & \textcolor{red}{-0.02} & \textcolor{blue}{0.00} & \textcolor{red}{-0.02} & \textcolor{red}{-0.03} & \textcolor{blue}{0.00} & \textcolor{red}{-0.03} & \textcolor{blue}{0.00} & \textcolor{blue}{+0.01} & \textcolor{red}{-0.01} & \textcolor{blue}{0.00} & \textcolor{blue}{+0.01} & \textcolor{red}{-0.02} & \textcolor{blue}{0.00} & \textcolor{red}{-0.01}\\

\hline

ES (Before) & 0.75 & 0.77 & 0.77 & 0.78 & 0.76 & 0.75 & 0.76 & 0.75 & 0.75 & 0.76 & 0.75 & \textbf{0.76} & 0.76 & 0.76 & 0.75 & 0.76\\
ES (After) & 0.75 & 0.76 & 0.77 & 0.75 & 0.76 & 0.76 & 0.76 & 0.75 & 0.75 & 0.76 & 0.75 & 0.77 & 0.74 & 0.76 & 0.73 & 0.75 \\

\textbf{Effect} & \textcolor{blue}{0.00} & \textcolor{red}{-0.01} & \textcolor{blue} {0.00} & \textcolor{red}{-0.03} & \textcolor{blue}{0.00} & \textcolor{blue}{+0.01} & \textcolor{blue}{0.00} & \textcolor{blue}{0.00} & \textcolor{blue}{0.00} & \textcolor{blue}{0.00} & \textcolor{blue}{0.00} & \textcolor{blue}{+0.01} & \textcolor{red}{-0.02} & \textcolor{blue}{0.00} & \textcolor{red}{-0.02} & \textcolor{red}{-0.01}\\

\hline
SE (Before) & 0.70 & 0.72 & 0.69 & 0.71 & 0.74 & 0.70 & 0.70 & 0.70 & 0.71 & 0.71 & 0.71 & 0.72 & \textbf{0.70} & 0.70 & 0.70 & 0.71\\
SE (After) & 0.69 & 0.72 & 0.71 & 0.71 & 0.71 & 0.71 & 0.69 & 0.70 & 0.68 & 0.70 & 0.69 & 0.71 & 0.70 & 0.68 & 0.69 & 0.70 \\

\textbf{Effect} & \textcolor{red}{-0.01} & \textcolor{blue}{0.00} & \textcolor{blue} {+0.02} & \textcolor{blue} {0.00} & \textcolor{red}{-0.03} & \textcolor{blue}{+0.01} & \textcolor{red}{-0.01} & \textcolor{blue}{0.00} & \textcolor{red}{-0.03} & \textcolor{red}{-0.01} & \textcolor{red}{-0.02} & \textcolor{red}{-0.01} & \textcolor{blue}{0.00} & \textcolor{red}{-0.02} & \textcolor{red}{-0.01} & \textcolor{red}{-0.01}\\

\hline

US (Before) & 0.70 & 0.71 & 0.71 & 0.71 & 0.73 & 0.73 & 0.71 & 0.71 & 0.70 & 0.73 & 0.70 & 0.71 & 0.71 & \textbf{0.72} & 0.72 & 0.71\\

US (After) & 0.72 & 0.73 & 0.73 & 0.72 & 0.71 & 0.72 & 0.71 & 0.74 & 0.71 & 0.73 & 0.70 & 0.72 & 0.71 & 0.73 & 0.72 & 0.72\\

\textbf{Effect} & \textcolor{blue}{+0.02} & \textcolor{blue}{+0.02} & \textcolor{blue} {+0.02} & \textcolor{blue}{+0.01} & \textcolor{red}{-0.02} & \textcolor{red}{-0.01} & \textcolor{blue}{0.00} & \textcolor{blue}{+0.03} & \textcolor{blue}{+0.01} & \textcolor{blue}{0.00} & \textcolor{blue}{0.00} & \textcolor{blue}{+0.01} & \textcolor{blue}{0.00} & \textcolor{blue}{+0.01} & \textcolor{blue}{0.00} & \textcolor{blue}{+0.01}\\

\hline

ZM (Before) & 0.68 & 0.69 & 0.69 & 0.70 & 0.71 & 0.67 & 0.67 & 0.67 & 0.69 & 0.68 & 0.67 & 0.70 & 0.69 & 0.69 & \textbf{0.68} & 0.69\\

ZM (After) & 0.69 & 0.68 & 0.68  & 0.67  & 0.69 & 0.68 & 0.69 & 0.69  & 0.69  & 0.69  & 0.68  & 0.68  & 0.69  & 0.69  & 0.69  & 0.69\\

\textbf{Effect} & \textcolor{blue}{+0.01} & \textcolor{red}{-0.01} & \textcolor{red}{-0.01} & \textcolor{red}{-0.03} & \textcolor{red}{-0.02} & \textcolor{blue}{+0.01} & \textcolor{blue}{+0.02} & \textcolor{blue}{+0.02} & \textcolor{blue}{0.00} & \textcolor{blue}{+0.01} & \textcolor{blue}{+0.01} & \textcolor{red}{-0.02} & \textcolor{blue}{0.00} & \textcolor{blue}{0.00} & \textcolor{blue}{+0.01} & \textcolor{blue}{0.00}\\

\hline
% the rest aren't complete yet

\end{tabular}
\caption{Effect of explicit generator country specification on emotion attribution accuracy for the DeepSeek model. Red values indicate performance decreases, while blue ones indicate performance increases. Note that AU = Australia, AT = Austria, BR = Brazil, BG = Bulgaria, CN = China, FI = Finland, HN = Honduras, IN = India, MW = Malawi, NL = Netherlands, NZ = New Zealand, ES = Spain, SE = Sweden, US = USA, ZM = Zambia.}
\label{tab:generator-context}
\end{table*}

\end{document}